\documentclass[pdflatex,sn-mathphys-num]{sn-jnl}
\usepackage{graphicx}
\usepackage{multirow}
\usepackage{amsmath,amssymb,amsfonts}
\usepackage{amsthm}
\usepackage{mathrsfs}
\usepackage[title]{appendix}
\usepackage{xcolor}
\usepackage{textcomp}
\usepackage{manyfoot}
\usepackage{booktabs}
\usepackage{algorithm}
\usepackage{algorithmicx}
\usepackage{algpseudocode}
\usepackage{listings}
\usepackage{subfig}
\usepackage{graphicx}

\usepackage{tikz}
\usetikzlibrary{shapes.geometric, arrows.meta, positioning}

\tikzset{
  startstop/.style = {rectangle, rounded corners, minimum width=3cm, minimum height=0.8cm, text centered, draw=black, fill=blue!10, font=\small},
  process/.style = {rectangle, minimum width=3cm, minimum height=1cm, text centered, draw=black, fill=gray!10, font=\small},
  decision/.style = {diamond, aspect=2, minimum width=3cm, minimum height=1cm, text centered, draw=black, fill=orange!30, font=\small},
  arrow/.style = {thick,->,>=stealth},
  yesno/.style = {midway, fill=white, font=\footnotesize, inner sep=1pt}
}

\theoremstyle{thmstyletwo}

\theoremstyle{thmstylethree}

\newcommand{\review}[1]{
    \ifdefined\ModeProduction
        {\color{black}#1}
    \else
        {\color{red}#1}
    \fi
}

\begin{document}

\title{\emph{e-Profits}: A Business-Aligned Evaluation Metric for Profit-Sensitive Customer Churn Prediction}

\author*[1,2]{\fnm{Awais} \sur{Manzoor}}\email{malik.awaismanzoor@gmail.com}
\affil*[1]{ADAPT Centre, eXplainable Analytics Group}
    
\author[1,2]{\fnm{M. Atif} \sur{Qureshi}} 
    
\author[2]{\fnm{Etain} \sur{Kidney}} 
\affil[2]{Faculty of Business, Technological University Dublin}

\author[3]{\fnm{Luca} \sur{Longo}} 

\affil[3]{
Department of Computer Science and Information Technology, University College Cork, Ireland}

\abstract{Retention campaigns in customer relationship management often rely on churn prediction models evaluated using traditional metrics such as AUC and F1-score. However, these metrics fail to reflect financial outcomes and may mislead strategic decisions. We introduce \emph{e-Profits}, a novel business-aligned evaluation metric that quantifies model performance based on customer lifetime value, retention probability, and intervention costs. Unlike existing profit-based metrics such as Expected Maximum Profit, which assume fixed population-level parameters, \emph{e-Profits} uses Kaplan-Meier survival analysis to estimate tenure-conditioned (customer-level) one-period retention probabilities and supports granular, per-customer profit evaluation. We benchmark six classifiers across two telecom datasets (IBM Telco and Maven Telecom) and demonstrate that \emph{e-Profits} reshapes model rankings compared to traditional metrics, revealing financial advantages in models previously overlooked by AUC or F1-score. The metric also enables segment-level insight into which models maximise return on investment for high-value customers. \emph{e-Profits} provides a transparent, customer-level evaluation framework that bridges predictive modelling and profit-driven decision-making in operational churn management. All source code is available at: \url{https://github.com/Awaismanzoor/eprofits}.}


\maketitle

\section{Introduction}
\label{Introduction}
Customer churn, the end of a relationship between a customer and a business, presents a serious risk to a company's sustainability. Whether customers choose to go to a competitor or stop using the service altogether, churn results in immediate revenue loss and a gradual decline in customer equity.\review{Since attracting new customers is often reported to be more expensive than keeping current ones, with many studies and managerial sources suggesting that acquisition costs substantially exceed retention costs \cite{van2004customer,zhou2023early}, many companies focus on customer retention strategies enhanced by predictive analytics
\cite{ganesh2000understanding,sun2021case}.}Loyal customers may gradually acquire more goods and services over time and are associated with reduced marketing costs \cite{loureiro2025improving}, and they spread positive sentiment, which indirectly attract new customers. Therefore, higher retention directly increases company profits.

Businesses increasingly rely on predictive models trained on behavioural and transactional data to address churn and identify customers at risk of leaving \cite{kotan2025novel,peng2023research}. Customer churn prediction (CPP) is typically framed as a binary classification task in the machine learning literature \cite{faraji2024customer}. The goal is to identify would-be churners early enough to intervene, usually via targeted retention campaigns offering incentives. However, while predictive accuracy is important, most models are evaluated using traditional metrics such as F1-score or AUC, which do not reflect real-world business impact \cite{hand2009measuring, elkan2001foundations}. This issue is especially problematic in profit-sensitive domains, where not all customers are equally valuable, and not all classification errors have the same cost \cite{bahnsen2015novel,thakkar2022clairvoyant}. For instance, targeting low-value customers who were unlikely to churn wastes retention resources, while failing to identify high-value churners results in substantial lost revenue. Studies have shown that an inverse relationship may exist between customer lifetime value (CLV) and churn probability \cite{rahman2025profit, lemmens2020managing}, meaning that models can systematically overlook high-value customers optimised for statistical accuracy.

To bridge this gap, profit-based evaluation metrics have been proposed, including the Maximum Profit Criterion (MPC) and Expected Maximum Profit (EMP) \cite{verbeke2012new, verbraken2012novel}. These metrics represent important progress but are often constrained by fixed assumptions, such as uniform CLV and constant offer acceptance probabilities, which rarely hold in real-world business contexts characterised by customer heterogeneity.

In this study, we introduce \emph{e-Profits}, a business-aligned evaluation metric designed to assess churn models based on financial relevance rather than statistical precision. By incorporating customer-specific lifetime value (CLV), intervention costs, and retention probabilities estimated via Kaplan-Meier survival analysis, \emph{e-Profits} enables granular, per-customer profitability assessment. We empirically evaluate six machine learning classifiers across two real-world telecom datasets (IBM Telco and Maven Telecom) and show that \emph{e-Profits} reveals model rankings that diverge significantly from those obtained using traditional predictive metrics and existing profit-based alternatives. Beyond technical robustness, \emph{e-Profits} aligns model evaluation with organisational decision-making by translating predictive outputs into financially actionable insights that are interpretable by both technical and non-technical stakeholders.

\review{

This paper introduces \emph{e-Profits}, a business-aligned evaluation metric for customer churn prediction that enables model selection based on financial impact rather than predictive accuracy alone. Unlike conventional metrics such as AUC, Accuracy, and F1-score, and unlike existing profit-based metrics that rely on restrictive population-level assumptions, \emph{e-Profits} provides a transparent, customer-level profit evaluation framework that can be applied to any trained classifier without modifying the learning algorithm. \\

The study makes the following key contributions:
\begin{enumerate}
    \item Introduce \emph{e-Profits}, a business-aligned evaluation metric for profit-sensitive churn prediction that integrates customer-level value, intervention costs, and retention probabilities;
    \item Incorporate survival-based retention estimation using both average (ARR) and tenure-conditioned (TRR) Kaplan-Meier formulations; 
    \item Demonstrate that profit-aware evaluation induces statistically significant model re-ranking relative to traditional objectives such as AUC and F1-score;
    \item Analyses both full-population and budget-constrained (top-segment) targeting scenarios relevant to real-world retention campaigns; and
    \item Validates the robustness of profit-based model selection using bootstrap confidence intervals and paired non-parametric tests across two real-world datasets.
\end{enumerate}
}

The remainder of the article is organised as follows. The next section \ref{sec:lit} reviews the relevant literature. Following that, section \ref{sec:methodology} discusses the methodology and experimental setup. Sections \ref{sec:results} and \ref{sec:discussion} present the results and a discussion of their implications, respectively. Finally, the section \ref{sec:conclusion} summarises the main findings of the work.

\section{Related Works}
\label{sec:lit}

Customer churn prediction has been extensively studied across domains such as telecommunications, banking, and subscription-based services \cite{bhattacharyya2021investigation}. Traditionally, churn is formulated as a supervised binary classification problem, with models trained to distinguish between customers who stay and those who leave. Early approaches employed logistic regression and decision trees, gradually evolving toward advanced techniques, including artificial neural networks \cite{mena2024exploiting}, support vector machines \cite{ly2022churn}, ensemble methods \cite{tavassoli2022hybrid}, and feature selection pipelines \cite{sana2022novel}. Despite these advances, the dominant focus remains on improving predictive performance measured through conventional metrics such as Accuracy, F1-score, and AUC \cite{stripling2018profit}.

These conventional metrics, however, are ill-suited for business settings where the cost of misclassification varies across customers. For example, AUC treats false positives and false negatives symmetrically \cite{hand2009measuring}, ignoring the financial impact of retention decisions. As Elkan noted in his foundational work on cost-sensitive learning \cite{elkan2001foundations}, a classifier’s utility must be assessed based on domain-specific cost matrices rather than classification error alone. In churn prediction, misclassifying a high-value customer as a non-churner could lead to significant revenue loss, while misclassifying a low-value, loyal customer could result in wasted marketing resources.

Profit-based evaluation metrics have been introduced to address this gap. Verbeke et al. introduced the Maximum Profit Criterion (MPC) \cite{verbeke2012new}, which estimates profit based on classification outcomes. This was extended by Verbraken et al. through the Expected Maximum Profit (EMP) metric \cite{verbraken2012novel}, which incorporates uncertainty in retention success and introduces the profit-maximising churn fraction, the optimal subset of customers to target. These metrics marked an important shift from accuracy-focused to business-aligned evaluation.\review{Other approaches focus on value-aware targeting and campaign evaluation under operational constraints \cite{rahman2025profit}. This approach propose fairness-aware preprocessing (e.g., reweighing and resampling) to improve representation and targeting high-CLV customers, and introduce expected profit-based evaluation across target fractions (the Area Under the Expected Profit Curve) to reflect budget-limited retention strategies. However, these value-aware approaches typically require modifying the training data or the learning objective (e.g., via reweighing/resampling), and their conclusions remain sensitive to how CLV segments and profit parameters are specified.}Table~\ref{tab:metric_comparison} summarises the key characteristics of traditional and profit-aware evaluation metrics, highlighting the business relevance and flexibility offered by \emph{e-Profits} compared to widely used alternatives.

Nevertheless, limitations persist. EMP assumes constant values for customer lifetime value (CLV), contact cost, and offer acceptance probability \cite{oskarsdottir2018profit}. It models retention success using an unimodal Beta distribution with fixed hyperparameters, which may not capture the heterogeneity of real-world customers. Additionally, while some studies have explored embedding profit functions directly into model optimisation via profit-aware decision trees or evolutionary strategies \cite{stripling2018profit,hoppner2020profit}.\review{For instance, ProfLogit replaces likelihood-based fitting with an EMP-based fitness function and uses evolutionary optimisation to search for model parameters that maximise expected campaign profit \cite{stripling2018profit}.} These approaches often introduce complexity and reduce interpretability, limiting their uptake in business environments.

Some recent works have considered time-dependent churn risk using survival analysis models \cite{perianez2016churn, xu2022lads}. These approaches explicitly demonstrate the value of incorporating tenure and retention time, but they seldom translate into actionable profit-based evaluation metrics.

\begin{table}[ht]
\centering
\setlength{\tabcolsep}{2.5pt}
\fontsize{8pt}{12pt}\selectfont
\renewcommand{\arraystretch}{1.1}
\caption{Comparison of churn evaluation metrics}
\label{tab:metric_comparison}

\begin{tabular}{p{2.5pc}p{4pc}p{3.5pc}p{4pc}p{4pc}p{6pc}p{6pc}}
\hline
\textbf{Metric} & \textbf{Type} & \textbf{Business-Aware} &
\textbf{Fixed Assumptions} & \textbf{Individual-Level} &
\textbf{Optimisation Target} & \textbf{Interpretability} \\
\hline
AUC & Ranking & No & -- & No & Churn likelihood &
High (threshold-free statistic) \\
F1-score & Classification & No & -- & No & Class balance &
High (standard classification metric) \\
EMP & Profit-based & Partial & Yes & No & Profit-max subset &
Medium (requires Beta and cost inputs) \\
AUEPC & Profit-based & Yes & Partial & \emph{Ranking-level only} & Area under profit curve &
Medium (campaign-level profit curve; customer heterogeneity enters via ordering) \\
\emph{e-Profits} & Profit-based & Yes & No & Yes & Total and segment profit &
High (transparent, per-customer profit decomposition) \\
\hline
\end{tabular}
\end{table}

While the academic literature contains several profit-aware metrics, their adoption in operational churn platforms remains rare. Surveys such as García et al. \cite{garcia2017intelligent} and our own review \cite{manzoor2024review} highlight that many practitioners still default to traditional metrics like AUC or lift, due to limitations in available tools and ease of understanding. This disconnect reinforces the need for practical and business-aligned evaluation metrics that can bridge the research-practice divide.

\review{Marketing research has long emphasized that churn behavior is heterogeneous across customers and that ignoring such heterogeneity can lead to poor managerial decisions \cite{fader2010customer}. In addition, churn risk is not equivalent to intervention responsiveness. Ascarza \cite{ascarza2018retention} highlights that targeting customers solely because they are high risk can be ineffective if those customers are unlikely to change their behaviour in response to retention actions. This distinction is important for our setting. We therefore position \emph{e-Profits} as a decision-oriented evaluation and model-selection criterion for a targeting policy: it maps a model’s churn scores to intervention decisions and aggregates the resulting costs and retained value under explicit business assumptions. Integrating heterogeneous responsiveness to interventions (e.g., uplift or treatment-effect modelling) is a complementary extension.}

\review{Despite these advances, existing profit-aware approaches often lack a simple and granular way to evaluate churn classifiers under customer-level heterogeneity in value, costs, and retention dynamics. Most rely on fixed cost parameters or make strong assumptions about offer acceptance rates. By contrast, our proposed \emph{e-Profits} is aimed at leveraging individual-level cost, revenue, and survival-based retention to provide a metric that is granular, understandable, and practically deployable, without retraining the model. \emph{e-Profits} responds to these limitations by offering a modular evaluation metric that aligns directly with business impact, empowering practitioners to make financially sound model selection decisions. In addition, we rigorously test whether resulting model re-ranking is statistically significant by reporting confidence intervals, non-parametric paired tests, and rank correlation analysis on held-out test data.}

\section{Methodology and Experimental Setup}
\label{sec:methodology}

This section introduces our proposed evaluation metric, \emph{e-Profits}, and explains how it aligns churn prediction with business objectives by incorporating customer-specific profitability and cost structure. We also introduce the details of a primary empirical study including the datasets selected, the classification techniques for churn prediction, and the comparative metrics for the purpose of validation.

\subsection{Overview and Motivation}

As established in the previous section, conventional evaluation metrics fail to capture the financial implications of churn misclassification, and existing profit-based metrics often rely on strong population-level assumptions. In response, we introduce \emph{e-Profits}, a modular evaluation metric that supports granular, business-aligned model comparison. The \emph{e-Profits} metric evaluates model performance by aggregating per-customer profit outcomes, combining customer-specific lifetime value, tenure-conditioned one-period retention probabilities derived via Kaplan-Meier survival analysis, and individualised intervention costs including contact and retention offers. This design enables profit-based comparison of churn models without altering training pipelines, ensuring both understandability and practical deployment feasibility.

\begin{figure}[!ht]
\centering

\begin{center} 
\begin{tikzpicture}[node distance=1.2cm and 2.2cm]

\tikzstyle{startstop} = [rectangle, rounded corners, minimum width=3.2cm, minimum height=1cm,text centered, draw=black, fill=gray!10]
\tikzstyle{process} = [rectangle, minimum width=3.2cm, minimum height=1cm, text centered, draw=black, fill=blue!10]
\tikzstyle{decision} = [diamond, aspect=2, text centered, draw=black, fill=green!15]
\tikzstyle{arrow} = [thick,->,>=stealth]
\tikzstyle{yesno} = [font=\small\itshape]

\node (start) [startstop] {Model Prediction};

\node (tpfp) [decision, below=0.6cm of start] {TP or FP?};

\node (zero) [process, right=1.0cm of tpfp] {e-Profits = 0};

\node (cost) [process, below=1.0cm of tpfp, align=left] 
{Cost of Intervention: \\ - Cost of Offer \\ - Cost of Contact};

\node (tpdec) [decision, below=1.0cm of cost] {True Positive?};

\node (clv) [process, below left=1.5cm and 0.8cm of tpdec, align=center]
{Calculate CLV \\ using eq. 3};

\node (tp) [process, below=0.8cm of clv, align=center] 
{e-Profits = \\ CLV - Cost of offer- Cost of contact};

\node (fp) [process, below right=1.6cm and 1.2cm of tpdec, align=center] 
{e-Profits = \\ - Cost of offer - Cost of contact};

\node (total) [startstop, below=5.5cm of tpdec, xshift=0cm] 
{Total e-Profits = $\sum$ e-Profits};

\draw [arrow] (start) -- (tpfp);
\draw [arrow] (tpfp.east) -- node[yesno, above] {No} (zero.west);

\draw [arrow] (tpfp.south) -- ++(0,-0.5) node[yesno, right, xshift=-0.2cm] {Yes} -- (cost.north);

\draw [arrow] (cost.south) -- ++(0,-0.5) -- (tpdec.north);

\draw [arrow] (tpdec.south) -- ++(0,-0.4) coordinate (split);

\draw [arrow] (split) -- ++(-2.4,0) node[yesno, above] {Yes} -| (clv.north);

\draw [arrow] (clv.south) -- (tp.north);

\draw [arrow] (split) -| node[yesno, above right, xshift=2pt] {No} (fp.north);

\draw [arrow] (tp.south) -- ++(0,-0.7) -| (total.north);
\draw [arrow] (fp.south) -- ++(0,-2.42) -| (total.north);

\end{tikzpicture}
\end{center}

\caption{Conceptual workflow of the \emph{e-Profits} evaluation metric. This flow illustrates how model predictions interact with customer-level retention, CLV, and cost information to compute aggregate business impact. (To calculate CLV, we use ARR and TRR as one-period retention probabilities derived from Kaplan-Meier.)}
\label{fig:eprofits-flow}
\end{figure}

\review{\subsection{\emph{e-Profits} formulation}
We define \emph{e-Profits} as the total estimated profit that would be realised if a retention campaign was executed based on the model’s churn predictions. Consider a dataset of \(N\) customers indexed by \(i = 1, \dots, N\). For each customer \(i\) we denote by \(y_i \in \{0,1\}\) the true churn label (1 indicates churn in the observation window), and by \(\hat{y}_i \in \{0,1\}\) the binary churn prediction produced by a classifier at a fixed decision horizon. The term \(\Pi_i\) represents the incremental profit or loss associated with customer \(i\) under a retention policy that targets customers with \(\hat{y}_i = 1\). The proposed overall \emph{e-Profits} is computed as the sum of individual customer-level profit contributions:

\begin{equation}
\label{eq:eprofits_sum}
\text{\emph{e-Profits}} = \sum_{i=1}^{N} \Pi_i.
\end{equation}

Eq. \ref{eq:eprofits_sum} states that \emph{e-Profits} is an aggregate measure obtained by adding up $\Pi_i$ which is the estimated profit or loss associated with customer $i$. We use a fixed threshold of 0.5 to obtain binary predictions $\hat{y}_i$ from predicted probabilities.

\subsubsection{Customer-Level Profit Function}

For each customer $i$, profit is determined by the interaction between the true churn outcome $y_i \in \{0,1\}$, the model prediction $\hat{y}_i \in \{0,1\}$, the customer lifetime value $\text{CLV}_i$, and the cost of intervention. The per-customer profit is formally defined as:

\begin{equation}
\Pi_i =
\begin{cases}
\text{CLV}_i - C_{\text{offer}, i} - C_{\text{contact}, i} & \text{if } y_i = 1 \text{ and } \hat{y}_i = 1 \quad \text{(True Positive)} \\\\
- C_{\text{offer}, i} - C_{\text{contact}, i} & \text{if } y_i = 0 \text{ and } \hat{y}_i = 1 \quad \text{(False Positive)} \\\\
0 & \text{otherwise (no retention action)}
\end{cases}
\label{eq:eprofits}
\end{equation}

This formulation reflects practical churn intervention logic: only customers predicted to churn receive retention actions. Each branch corresponds to one of the four possible combinations of prediction and true label for customer \(i\). $\text{CLV}_i$ is the estimated customer lifetime value, and  $C_{\text{offer}, i}$ and $C_{\text{contact}, i}$ are the intervention costs for customer $i$. When the model predicts churn (\(\hat{y}_i = 1\)) and the customer is in fact a churner (\(y_i = 1\)), we assume that the intervention prevents churn and the firm retains the estimated customer lifetime value \(\text{CLV}_i\) but pays the offer and contact costs. This gives \(\Pi_i = \text{CLV}_i - C_{\text{offer},i} - C_{\text{contact},i}\) for true positives. When the model predicts churn but the customer would not churn (\(y_i = 0\)), the intervention does not increase revenue, yet the firm still incurs the offer and contact costs. The incremental profit is therefore negative and equal to \(- C_{\text{offer},i} - C_{\text{contact},i}\) for false positives. For false negatives and true negatives (\(\hat{y}_i = 0\)), no action is taken and no retention cost is incurred. Since \emph{e-Profits} evaluates the incremental profit of the decisions made under the targeting policy, we assign \(\Pi_i = 0\) in these cases. This makes eq. \ref{eq:eprofits} a transparent mapping from classification outcomes to profit contributions. We emphasise that \emph{e-Profits} evaluates the profitability of a targeting policy under explicit cost and retention assumptions; it does not estimate the causal treatment effect of an intervention.

\subsubsection{Customer Lifetime Value (CLV)}

Customer lifetime value for customer \(i\) is computed by combining their revenue, a fixed profit margin and estimated retention probability:

\begin{equation}
\label{eq:clv}
\text{CLV}_i = \frac{R_i \times M}{1 - \min(r_i, 0.995)}
\end{equation}

Where $R_i$ denotes the monthly revenue of customer $i$, $M$ is a fixed profit margin, and $r_i$ is the estimated one-period retention probability (probability of remaining active for one additional billing period). The denominator $(1-r_i)$ corresponds to the churn probability in a geometric-like revenue model, yielding the familiar perpetuity form commonly used in marketing and finance. Retention probabilities $r_i$ are obtained from a Kaplan-Meier survival estimator as a function of customer tenure, prioritising simplicity, low-overhead and transparency. To avoid unrealistically large CLV values when estimated retention probabilities approach one, we cap $r_i$ at 0.995 as a numerical safeguard; this cap does not represent a behavioural assumption. We emphasise that \emph{e-Profits} evaluates the profitability of a targeting policy under explicit cost and retention assumptions and does not estimate causal treatment effects.
}

In this framework, retention actions are only initiated when $\hat{y}_i = 1$ (i.e., the model predicts churn). True positives generate profit, while false positives incur unnecessary cost. No intervention, and hence no financial consequence, is applied to customers predicted as non-churners.

\review{\subsubsection{Intervention Cost Model}

Intervention cost comprises two components: the cost of making contact with the customer and the cost of a retention offer. Both are proportional to the customer's CLV.

\paragraph{Retention Offer Cost}
\begin{equation}
\label{eq:offer_cost}
C_{\text{offer}, i} = \text{CPO} \cdot \text{CLV}_i
\end{equation}

where $\text{CPO}$ denotes Cost-per-offer ratio (e.g., 0.1).

\paragraph{Customer Contact Cost}

\begin{equation}
\label{eq:contact_cost}
C_{\text{contact}, i} = \max \left( c_0,\; c_1 \cdot C_{\text{offer}, i} \right)
\end{equation}

where $c_0$ represents a minimum fixed contact cost and $c_1$ scales variable contact cost proportionally to the offer cost. This parametrisation reflects the intuition that high value customers tend to receive more expensive offers and may be contacted through more resource intensive channels, while keeping the model simple and easy to calibrate. This structure reflects realistic business conditions where retention cost increases with customer value, while respecting minimum outreach overhead. Complete workflow of \emph{e-Profits} is illustrated in figure \ref{fig:eprofits-flow}.

\subsubsection{Retention Probability Estimation via Survival Analysis}

To estimate the probability that a customer remains subscribed beyond a given tenure, we employ Kaplan-Meier survival analysis \cite{perianez2016churn}. This method is non-parametric and does not assume a specific distribution of churn times. In our setting, retention varies at the customer level only through each customer's observed tenure using a single population-level survival curve. We therefore describe these quantities as tenure-based retention probabilities. The term tenure-conditioned retention emphasise that retention varies across customers only through current tenure $T_i$ via a single population Kaplan-Meier curve, rather than through a covariate-based individual survival model. We use the terms tenure-based retention and tenure-conditioned retention interchangeably to refer to retention probabilities derived from conditioning on current tenure via the Kaplan-Meier curve.

\noindent\textbf{Train/test separation and censoring}
We fit the Kaplan-Meier survival curve $S(\cdot)$ using only the training split to avoid any information leakage into model evaluation. Churn events are treated as failures, and customers who have not churned by the end of observation are treated as right-censored at their observed tenure, under the standard non-informative censoring assumption. The time origin is subscription start (onboarding), tenure is measured in months. The held-out test split is never used to fit $S(\cdot)$.

We consider two forms of retention probability:}

\begin{itemize}
    \review{\item \textbf{\textit{Tenure-based Retention Probability (TRR):}} A customer-level retention estimate that varies with the customer's current tenure $T_i$, computed from a population Kaplan-Meier survival curve. Let $T$ denote the customer lifetime random variable and let $S(t)$ be the Kaplan-Meier survival function. For a customer $i$ with current tenure $T_i$, we define the TRR as the one-period conditional survival probability over horizon $\Delta$:

\begin{equation}
\label{eq:trr}
r_i^{\text{TRR}}(\Delta)
= \Pr(T > T_i + \Delta \mid T > T_i)
= \frac{S(T_i + \Delta)}{S(T_i)}, \qquad i=1,\dots,N.
\end{equation}

In our experiments we set $\Delta=1$ month to align retention with the monthly revenue definition used in Eq.~\ref{eq:clv}. This definition explicitly conditions on the customer having already survived up to tenure $T_i$, and therefore matches the intended meaning of conditioned on current tenure.

    \item \textbf{\textit{Average Retention Probability (ARR):}} The average retention Probability is defined as the mean conditional survival probability over the empirical tenure distribution:

\begin{equation}
\label{eq:arr}
r^{\text{ARR}}(\Delta) = \frac{1}{N}\sum_{i=1}^{N} r_i^{\text{TRR}}(\Delta)
= \frac{1}{N}\sum_{i=1}^{N}\frac{S(T_i+\Delta)}{S(T_i)}.
\end{equation}

Equivalently, in continuous form:
\begin{equation}
r^{\text{ARR}}(\Delta) = \int_{0}^{\infty} \frac{S(t+\Delta)}{S(t)}\, dF(t),
\end{equation}}

\end{itemize}

\review{where $F(t)$ is the empirical distribution of observed customer tenures. The CLV formula uses both ARR and TRR to compute expected profitability. To prevent unrealistically large CLV values due to extremely high retention estimates, we cap the retention probability at a maximum value of 0.995, ensuring numerical stability in the denominator, as discussed in Eq. \ref{eq:clv}. Kaplan-Meier survival curves used to compute retention inputs (TRR/ARR) were fitted using only the training split. The held-out test split was never used to fit the survival curve or compute retention. 

During hyperparameter selection, we perform 5-fold cross-validation on the training split only. For each tuning target (AUC, Accuracy, F1, EMP, or \emph{e-Profits}), the corresponding metric is computed on validation folds using retention probabilities derived from a Kaplan-Meier survival curve $S(\cdot)$ that is estimated once on the training split and kept fixed throughout tuning. The Kaplan-Meier curve is therefore not refitted per fold. After hyperparameter selection, the chosen model is refit on the full training split and evaluated once on the held-out test split, which is never used to estimate retention.}

\review{\begin{algorithm}[t]
\caption{Retention-probability computation from a Kaplan-Meier curve}
\begin{algorithmic}[1]
\State Fit Kaplan-Meier survival curve $S(\cdot)$ using training churn times and censoring indicators
\For{each customer $i$}
    \State Observe current tenure $T_i$
    \State Compute $r_i^{\text{TRR}}(\Delta) \leftarrow \frac{S(T_i+\Delta)}{S(T_i)}$
\EndFor
\State Compute $r^{\text{ARR}}(\Delta) \leftarrow \frac{1}{N}\sum_{i=1}^{N} r_i^{\text{TRR}}(\Delta)$
\end{algorithmic}
\end{algorithm}
}

\subsection{Comparison with EMP and Implementation Guidance}

While EMP also aims to optimise profit, it differs from \emph{e-Profits} in several important respects. EMP assumes fixed customer lifetime value and global retention parameters, whereas \emph{e-Profits} operates at the customer level. EMP models offer acceptance via a parametric Beta distribution, while \emph{e-Profits} relies on empirically estimated retention probabilities obtained through survival analysis. Finally, EMP is analytically optimised for a specific target fraction, whereas \emph{e-Profits} enables direct computation of both total and segment-level profit without imposing restrictive population-level assumptions. These differences make \emph{e-Profits} more transparent and better suited for operational decision-making. These differences are summarised empirically in Table \ref{table:IBM} and \ref{table:MAVEN}.

The \emph{e-Profits} metric can be applied to any churn classifier without modifying the model architecture. Its implementation requires predicted churn labels, customer-level revenue and tenure information to compute lifetime value, retention probability estimation via Kaplan-Meier survival analysis, and specification of contact and offer cost parameters. This modular design allows \emph{e-Profits} to integrate seamlessly into standard evaluation pipelines and business dashboards used by marketing and 
customer analytics teams.

\review{\subsection{Computational overhead of Kaplan–Meier}
\label{overhead}
Kaplan-Meier estimation requires sorting observed churn/censoring times and computing a stepwise survival curve, with computational cost on the order of \(O(n\log n)\) due to sorting. The Kaplan-Meier curve is estimated once using the training split only (to avoid leakage) and then reused to derive retention inputs (ARR/TRR) throughout hyperparameter tuning and test evaluation; hence, KM estimation adds only a small, one-time preprocessing cost. During tuning, the dominant runtime cost arises from repeatedly fitting models and evaluating fold scores across hyperparameter configurations. 

Empirically, on the Maven dataset (mean over models), tuning on AUC required 147.48\,s, while tuning on \emph{e-Profits} required 150.12\,s (ARR) and 103.50\,s (TRR); EMP tuning required 287.50\,s. These timings (measured on the Maven dataset) indicate that incorporating Kaplan-Meier derived retention into \emph{e-Profits} does not add prohibitive overhead relative to standard objectives; the overall runtime is primarily governed by model training and the size of the hyperparameter grid.}

\subsection{Datasets}

We apply the \emph{e-Profits} metric using two publicly available real-world telecom churn datasets selected for their widespread use in recent research, public accessibility for reproducibility, and moderate classification difficulty (accuracy or F1-score typically \textless 85\%).

\textbf{1. IBM Telco Dataset}\footnote{\url{https://www.kaggle.com/datasets/blastchar/telco-customer-churn/data}}  
This dataset contains 7,043 customer records from a telecommunications provider, with 19 predictive features including customer demographics, tenure, services subscribed, and payment methods. The churn rate is 26.5\%.

\textbf{2. Maven Telecom Dataset}\footnote{\url{https://www.kaggle.com/datasets/johnp47/maven-churn-dataset}}  
This dataset includes 7,043 entries and 32 features. It presents a slightly different class distribution: 27\% churners, 67\% non-churners, and 6\% recent joiners. We excluded recent joiners for binary classification. Both datasets contain no missing values and require minimal preprocessing.
These datasets were selected to evaluate how \emph{e-Profits} performs in settings with moderate class imbalance and diverse customer features.

\subsection{Machine Learning Models}

A prior systematic review of churn prediction techniques \cite{manzoor2024review}, tree-based ensemble data-driven learning techniques have consistently demonstrated superior performance in predictive accuracy and robustness across various datasets. In particular, gradient-boosted techniques such as XGBoost, LightGBM, and CatBoost were frequently reported as top performers in learning accurate models, while Random Forest remained a strong baseline. Moreover, Explainable Boosting Machines were noted for offering a compelling balance between model accuracy and interpretability, an increasingly important consideration in decision-support applications \cite{liu2023concrete}.
In this study, we selected the following learning techniques for evaluation:

    \textbf{Random Forest (RF)}: Bagging-based ensemble proved to handle overfitting and high-dimensional data well \cite{bentejac2021comparative}.\\
    \indent\textbf{Gradient Boosting Classifier (GBC)}: A boosting algorithm that sequentially reduces residual errors \cite{alshourbaji2023efficient}.\\
    \indent\textbf{XGBoost (XGB)}: An optimised gradient boosting learning technique known for scalability and handling regularisation well \cite{paliwal2022xgbrs}.\\
    \indent\textbf{LightGBM (LGBM)}: A leaf-wise boosting algorithm with histogram-based split search optimal for accelerated training \cite{sina2022model}.\\
    \indent\textbf{CatBoost (CBC)}: a technique that can handle categorical variables natively and reduces prediction shift via ordered boosting \cite{asif2025data}.\\
    \indent\textbf{Explainable Boosting Machine (EBM)}: A generalised additive model with competitive performance and enhanced interpretability \cite{amekoe2024exploring}.

All models were trained using a 5-fold cross-validation strategy since the datasets do not contain many records. Table~\ref{tab:hyperparams} shows the hyperparameters and their ranges. This ensured fair performance comparison while accounting for model-specific tuning characteristics.

\begin{table}[ht]
\setlength{\tabcolsep}{3.2pt}
\fontsize{8pt}{12pt}\selectfont
    \renewcommand{\arraystretch}{1.1}
    \qquad
\centering
\caption{Hyperparameter tuning grids for each classifier}
\label{tab:hyperparams}
\begin{tabular}{p{1cm} p{11.5cm}}
\hline
\textbf{Model} & \textbf{Hyperparameters} \\
\hline
RF & \texttt{n\_estimators} = [100, 200, 300]; \texttt{max\_depth} = [None, 10, 20, 30]; \texttt{min\_samples\_split} = [2, 5, 10]; \texttt{min\_samples\_leaf} = [1, 2, 4] \\
XGB & \texttt{n\_estimators} = [100, 200, 300]; \texttt{learning\_rate} = [0.01, 0.1, 0.2]; \texttt{max\_depth} = [3, 4, 5]; \texttt{colsample\_bytree} = [0.3, 0.7] \\
GBC & \texttt{n\_estimators} = [100, 200, 300]; \texttt{learning\_rate} = [0.01, 0.1, 0.2]; \texttt{max\_depth} = [3, 4, 5] \\
LGBM & \texttt{n\_estimators} = [100, 200]; \texttt{learning\_rate} = [0.01, 0.1]; \texttt{num\_leaves} = [31, 50] \\
CBC & \texttt{iterations} = [100, 200]; \texttt{learning\_rate} = [0.01, 0.1]; \texttt{depth} = [3, 5] \\
EBM & \texttt{learning\_rate} = [0.01, 0.05, 0.1]; \texttt{max\_bins} = [128, 256]; \texttt{interactions} = [10, 20]; \texttt{max\_interaction\_bins} = [16, 32] \\
\hline
\end{tabular}
\end{table}

\subsection{Evaluation Metrics}

We benchmark \emph{e-Profits} against both traditional and profit-oriented evaluation metrics, including:\\
    \indent \textit{Accuracy:} Proportion of correctly classified instances. It is a simple metric, but it ignores class imbalance.\\
    \indent \textit{F1-score:} Harmonic mean of precision and recall, appropriate for better handling imbalanced binary classification.\\
    \indent \textit{AUC (Area Under the ROC Curve):} Measures ranking performance independent of decision threshold.\\
    \indent \textit{Top-Decile Lift:} Measures how concentrated churners are in the top 10\% of predicted probabilities.\\
    \indent \textit{Lift Index:} Weighted index of churners across ranked deciles; values close to 1 indicate effective prioritisation.\\
    \indent \textit{Expected Maximum Profit (EMP):} A profit-based metric assuming fixed CLV and offer acceptance probabilities \cite{verbraken2012novel}.\\
    \indent \textit{\emph{e-Profits} (ours):} Our proposed metric that evaluates models based on customer-specific CLV, retention, and intervention cost.\\

Metrics are reported in two scenarios: the \textit{full population evaluation}, which assesses the total profit across all predicted churners, and the \textit{top-segment evaluation}, focusing on the profit generated from the top 20\% of the most profitable customers based on predicted risk. This dual perspective allows us to understand both global model performance and targeted campaign efficiency\footnote{Further mathematical formulations for traditional evaluation metrics can be found in our earlier review work \cite{manzoor2024review}}.

From a deployment perspective, the \emph{e-Profits} metric is practically employable within firms. It integrates seamlessly with existing modelling workflows and provides direct financial interpretation that supports resource allocation decisions in marketing and sales. This makes it a high-utility tool for internal stakeholders responsible for campaign design and retention optimization.

\subsection{Model Selection and Validation Procedure}
\label{mdlselection}
We split the dataset once into a training set and a held-out test set. Hyperparameters are selected using 5-fold cross-validation on the training set only. For each tuning target (AUC, accuracy, F1, EMP, or e-Profits), the corresponding metric is used as the cross-validation score for grid search and to select the best hyperparameters; the final model is then refit on the full training set using the selected configuration. All reported results provided in Tables \ref{table:IBM} and \ref{table:MAVEN} are computed on the held-out test set, which is not used during tuning or model selection.

\section{Experimental Results}
\label{sec:results}

This section presents and analyses the performance of several machine learning classifiers for churn prediction using the two benchmark datasets: IBM Telco and Maven Telecom. Each classifier is evaluated using both traditional metrics (e.g., Accuracy, F1-score, AUC) and business-aligned metrics (e.g., top-decile lift, Lift Index, EMP, and the proposed \emph{e-Profits}). We aim to assess how model rankings and decision quality shift when profitability is considered alongside classification performance.

Results are organised in two parts. First, we report and discuss the results for the IBM Telco dataset, followed by the Maven dataset. We examine both full-population and top-segment outcomes for each dataset, focusing on how different tuning objectives (e.g., optimising for AUC vs. \emph{e-Profits}) influence downstream profit impact.

\subsection{IBM Dataset Analysis}

\begin{figure*}[!ht]
    \centering
    \subfloat[Kaplan-Meier estimated retention curve \label{fig:retention}]   {\includegraphics[height=50mm, width=65mm]{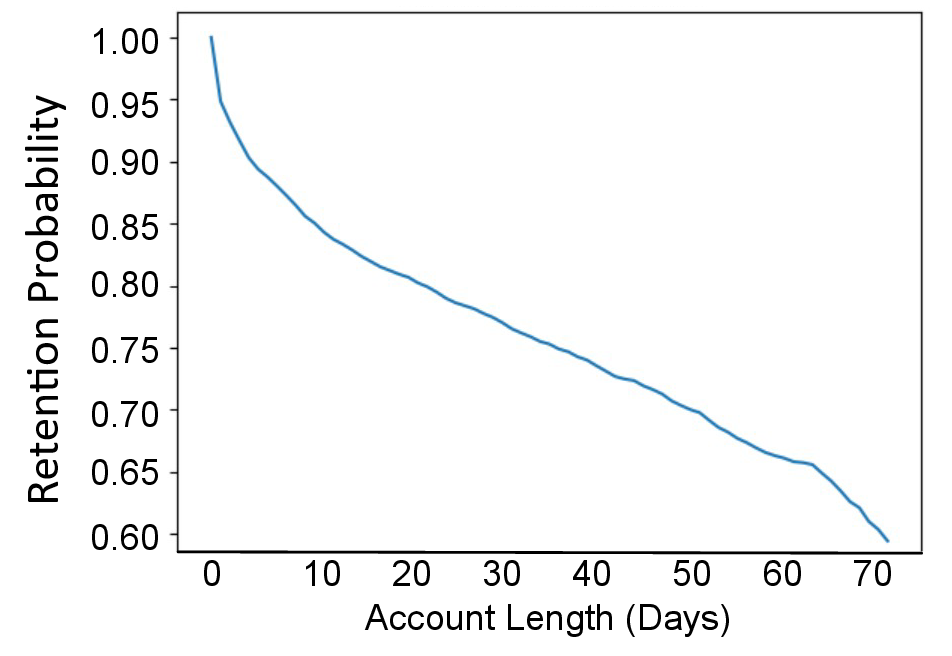}} \hfill
    \subfloat[Lift Index chart using EBM \label{fig:lift-IBM}]
    {\includegraphics[height=50mm, width=65mm]{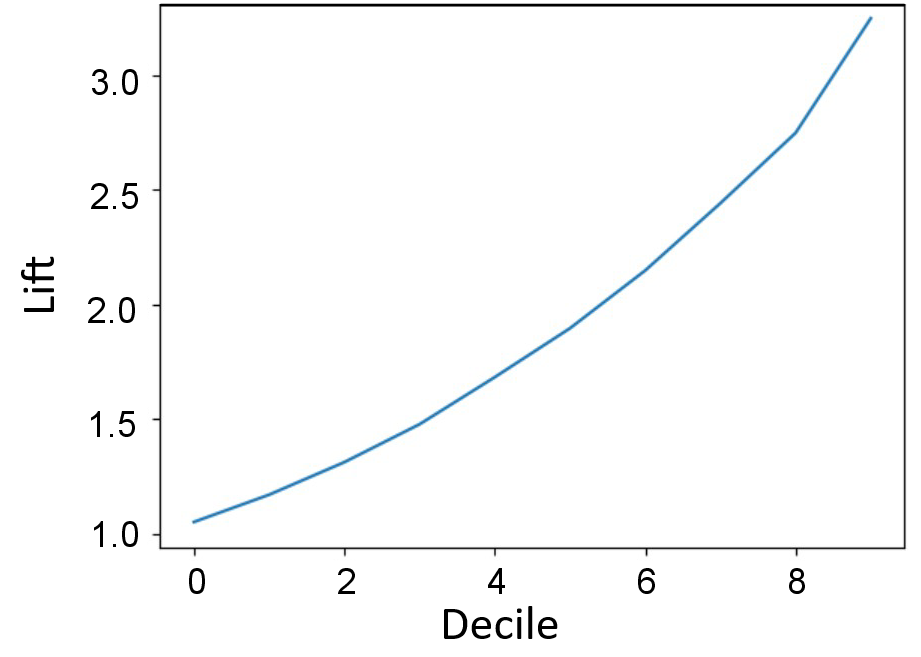}}   \hfill
    \subfloat[ROC curves (tuned on \emph{e-Profits}) \label{fig:AUC-IBM}]
    {\includegraphics[height=50mm, width=80mm]{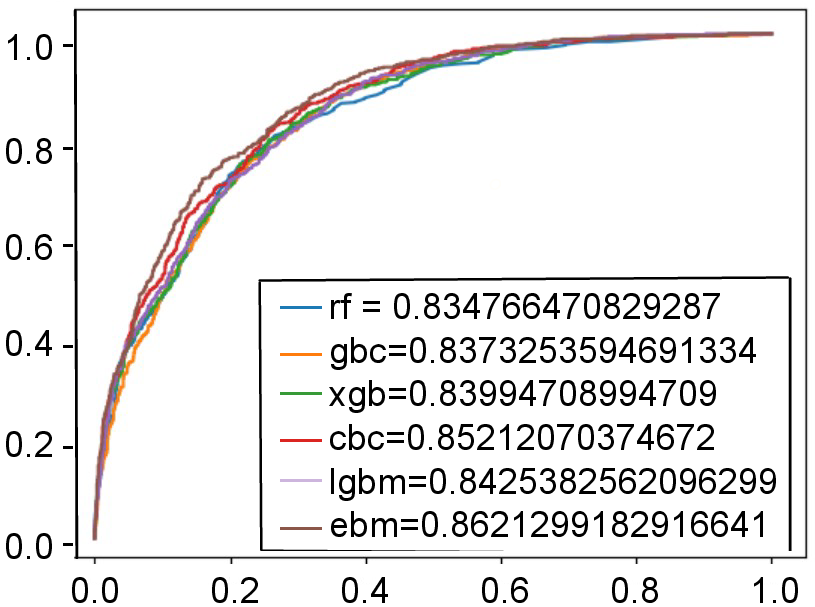}}   \hfill
    \caption{Overview of IBM dataset analysis: retention estimation, churn prioritisation, and classification performance.}
    \label{fig:ibm-overview}
\end{figure*}

We begin with the IBM Telco dataset, which contains customer tenure and service attributes relevant for churn prediction. Prior to model evaluation, we applied Kaplan-Meier survival analysis to estimate customer retention probabilities. The ARR estimated from the training split of the IBM dataset is approximately 0.77. This retention estimate is then treated as fixed and applied when evaluating models on the held-out test split. Figure~\ref{fig:ibm-overview} provides a compact overview, combining the retention curve, lift index, and ROC curve for visual comparison.

To assess how well models prioritise high-risk churners, we computed the lift index for the best-performing model under default AUC tuning (EBM). The top-decile lift of 3.01 indicates that the top 10\% of customers identified by the model are approximately three times more likely to churn than average, supporting targeted interventions. Similarly, ROC curves under \emph{e-Profits} tuning (Figure~\ref{fig:AUC-IBM}) show that EBM delivers the best AUC, reaffirming its strong discrimination ability.

\subsubsection{Classifier Performance across Metrics}

Table~\ref{table:IBM} reports the evaluation outcomes for different models under five tuning objectives: AUC, Accuracy, F1-score, EMP, and \emph{e-Profits}, respectively. For each setting, we report both full-population and top-segment \emph{e-Profits}, as well as corresponding EMP, Lift, and classification metrics.

To enrich our analysis of model profitability, we report \emph{e-Profits} for both the full population and the top 20\% most profitable customers, under two retention assumptions: the global Average Retention Rate (ARR) and the customer-level Tenure-based Retention Rate (TRR). Both are derived via Kaplan-Meier survival analysis, as described in the Methodology (see Section~\ref{sec:methodology}). This dual reporting allows us to examine how tenure-conditioned retention estimates affect model rankings under business-aligned metrics.

When optimised on AUC, EBM outperforms other classifiers on traditional predictive metrics, including AUC, Accuracy, F1-score, Top-decile lift, and EMP. However, Random Forest achieves higher \emph{e-Profits} on both the full-population and the top 20\% segment, indicating that AUC-optimised models do not necessarily prioritise customers with the highest financial value. While EBM consistently dominates classification performance across tuning objectives, models such as XGB and LGBM frequently achieve higher financial returns when evaluated using \emph{e-Profits}.

Examining profit outcomes in more detail reveals systematic differences between full-population and targeted evaluation. LGBM achieves the highest top-segment (20\%) \emph{e-Profits} under both ARR and TRR when tuned on F1-score, EMP, and \emph{e-Profits}. In contrast, for full-population profitability, XGB yields the highest \emph{e-Profits} when tuned on Accuracy, F1-score, and \emph{e-Profits}. When tuned on EMP, EBM attains a slightly higher full-population \emph{e-Profits} than XGB, with XGB ranked second. These results show that profit-based model rankings differ substantially from those induced by traditional predictive objectives. Crucially, models tuned directly on \emph{e-Profits} achieve the highest or near-highest realised profit across both full-population and top-segment evaluations, while maintaining competitive predictive performance.

Although EMP reflects a profit-oriented objective, its reliance on fixed cost and acceptance-rate assumptions limits its alignment with individual-level profit signals. As a result, models tuned using EMP do not consistently maximise \emph{e-Profits}; in particular, several models yield zero profit in the top segment under EMP tuning, an artefact of global CLV and offer-response assumptions rather than model capability. When models are tuned directly on \emph{e-Profits}, XGB achieves the highest overall profit, while LGBM and CBC yield superior outcomes for the top 20\% most profitable customers, as also illustrated by the radar plot in Figure~\ref{fig:Radar-IBM}.

\begin{table}[!htbp]
\centering
\caption{\review{Classifiers' performance on the IBM dataset under AUC, Accuracy, F1-score, EMP and \emph{e-Profits}, respectively.}}
\label{table:IBM}
\setlength{\tabcolsep}{3.2pt}
\fontsize{8pt}{12pt}\selectfont
\renewcommand{\arraystretch}{0.95}

\begin{tabular}{p{0.8cm} p{0.7cm} p{0.7cm} p{0.7cm} p{0.8cm} p{0.85cm}p{1.1cm} p{1.1cm} p{1.1cm} p{1.1cm} p{1.1cm} p{1.1cm}} 
\hline
\textbf{Name} & \textbf{F1 (\%)} & \textbf{Acc. (\%)} & \textbf{AUC (\%)} & \textbf{TDL} & \textbf{Lift Index} &
\textbf{e-Prof (0.2, 0.3, ARR)} & \textbf{e-Prof (0.2, 0.3, TRR)} &
\textbf{e-Prof (1.0, 0.3, ARR)} & \textbf{e-Prof (1.0, 0.3, TRR)} &
\textbf{EMP (0.2, 200, 10, 1)} & \textbf{EMP (1.0, 200, 10, 1)} \\
\hline

\multicolumn{12}{c}{\textbf{Tuned on AUC}} \\
\hline
RF    & 57.20 & 79.18 & 83.77 & 2.8961 & 0.7932 & \textbf{878821} & \textbf{744024} & \textbf{7289696} & \textbf{9104621} & 33.61 & 10.52 \\
GBC   & 57.82 & 76.72 & 83.82 & 2.6867 & 0.7951 & 0      & 0      & 852922  & 448270  & 32.34 & 10.65 \\
XGB   & 59.26 & 75.20 & 85.62 & 2.7216 & 0.8064 & 0      & 0      & 330130  & 151294  & 34.57 & 10.97 \\
CBC   & 57.82 & 77.99 & 84.85 & 2.7216 & 0.8031 & 0      & 0      & 2311538 & 1957368 & 33.62 & 10.76 \\
LGBM  & 56.25 & 79.03 & 84.72 & 2.9659 & 0.8003 & 0      & 0      & 2150792 & 7944468 & 33.61 & 10.67 \\
EBM   & \textbf{59.62} & \textbf{80.83} & \textbf{86.16} & \textbf{3.0008} & \textbf{0.8103} & 478040 & 701949 & 6675152 & 7894896 & \textbf{35.89} & \textbf{10.98} \\
\hline

\multicolumn{12}{c}{\textbf{Tuned on Accuracy}} \\
\hline
RF    & 57.20 & 80.03 & 85.64 & 3.0008 & 0.8075 & 136351 & 199609 & 5745843 & 6891397 & 34.91 & 10.83 \\
GBC   & 58.56 & 79.84 & 85.38 & 2.8787 & 0.8056 & 268062 & 511788 & 6792305 & 8518455 & 34.59 & 10.84 \\
XGB   & 59.32 & 80.27 & 85.48 & 2.8612 & 0.8051 & \textbf{608452} & \textbf{922701} & \textbf{7787442} & \textbf{9953379} & 34.57 & 10.87 \\
CBC   & 59.52 & 80.69 & 85.74 & 3.0008 & 0.8080 & 399446 & 614876 & 7059167 & 9283514 & 35.41 & 10.85 \\
LGBM  & 55.04 & 79.51 & 85.25 & 3.0182 & 0.8038 & 574879 & 855865 & 5741153 & 6941722 & 33.55 & 10.74 \\
EBM   & \textbf{60.61} & \textbf{81.12} & \textbf{86.15} & \textbf{3.0182} & \textbf{0.8118} & 565626 & 823097 & 7401027 & 8898532 & \textbf{36.15} & \textbf{10.93} \\
\hline

\multicolumn{12}{c}{\textbf{Tuned on F1-score}} \\
\hline
RF    & 57.20 & 80.03 & 85.64 & 3.0008 & 0.8075 & 136351 & 199609 & 5745843 & 6891397 & 34.91 & 10.83 \\
GBC   & 58.56 & 79.84 & 85.38 & 2.8787 & 0.8056 & 268062 & 511788 & 6792305 & 8518455 & 34.59 & 10.84 \\
XGB   & 59.51 & 80.36 & 85.58 & 2.9310 & 0.8071 & 668892 & 894586 & \textbf{7805429} & \textbf{9953379} & 34.96 & 10.86 \\
CBC   & 57.82 & 79.70 & 85.21 & 2.9833 & 0.8045 & 754358 & 1060291 & 7365250 & 9193007 & 34.41 & 10.77 \\
LGBM  & 56.25 & 78.80 & 84.25 & 2.9659 & 0.7981 & \textbf{1148846} & \textbf{1583632} & 6746884 & 8421249 & 33.35 & 10.65 \\
EBM   & \textbf{60.61} & \textbf{81.12} & \textbf{86.15} & \textbf{3.0182} & \textbf{0.8118} & 608452 & 922701 & 7787442 & 9833881 & \textbf{36.15} & \textbf{10.93} \\
\hline

\multicolumn{12}{c}{\textbf{Tuned on EMP}} \\
\hline
RF   & 51.11 & 79.08 & 85.43 & 2.8438 & 0.8052 & 0     & 0      & 322641 & 3204003 & 34.43 & 10.81 \\
GBC   & 56.09 & 79.70 & 85.27 & 2.8787 & 0.8051 & 10062 & 81201 & 5411529 & 6497417 & 34.09 & 10.79 \\
XGB   & 59.52 & 80.69 & 85.74 & 3.0008 & 0.8080 & 399446 & 614876 & 7701027 & 9783514 & 35.41 & 10.85 \\
CBC    & 57.67 & 80.27 & 85.65 & 3.0008 & 0.8085 & 136351 & 199609 & 5793988 & 6956820 & 35.04 & 10.85 \\
LGBM  & 56.87 & 79.60 & 84.68 & 3.0531 & 0.7998 & \textbf{686595} & \textbf{1069797} & 5784574 & 7082188 & 33.93 & 10.64 \\
EBM   & \textbf{60.84} & \textbf{81.12} & \textbf{86.22} &  \textbf{2.9833} & \textbf{0.8124} & 573412 & 889289 & \textbf{7732943} & \textbf{9829970} & \textbf{35.70} & \textbf{10.95} \\
\hline

\multicolumn{12}{c}{\textbf{Tuned on e-Profits (under TRR on full population)}} \\
\hline
RF    & 54.10 & 78.56 & 83.47 & 2.8438 & 0.7918 & 621741 & 584309 & 8776019 & 9150798 & 32.05 & 10.42 \\
GBC   & 56.53 & 78.89 & 83.73 & 2.7565 & 0.7936 & 1149186 & 1426809 & 7676514 & 9632075 & 32.19 & 10.60 \\
XGB   & 58.47 & 79.22 & 83.82 & 2.8612 & 0.7937 & 1014112 & 1240208 & \textbf{8312032} & \textbf{10991460} & 32.69 & 10.61 \\
CBC   & 57.82 & 79.70 & 85.21 & 2.9833 & 0.8045 & 1148846 & 1583632 & 7805429 & 9933881 & 34.41 & 10.77 \\
LGBM  & 57.09 & 78.66 & 83.50 & 2.8961 & 0.7913 & \textbf{1571624} & \textbf{1627728}  & 7530726 & 9564945 & 32.74 & 10.48 \\
EBM   & \textbf{59.05} & \textbf{80.50} & \textbf{86.09} &  \textbf{2.966} & \textbf{0.8110} & 552421 & 851399 & 7339760 & 9785358 & \textbf{36.05} & \textbf{10.94} \\
\hline

\end{tabular}

\begin{flushleft}
\footnotesize{*\emph{e-Profits} are reported as $e\text{-}Prof(\eta,M,\mathrm{RR})$, where $\eta\in\{0.2,1.0\}$ denotes the evaluated customer fraction (top 20\% or full population), $M=0.3$ is the profit margin used in Eq.~\ref{eq:clv}, and $\mathrm{RR}\in\{\mathrm{ARR},\mathrm{TRR}\}$ specifies whether average or tenure-based retention probabilities are used. EMP values are computed as $\mathrm{EMP}(\eta,\mathrm{CLV},C_{\text{offer}},C_{\text{contact}})$ with $\mathrm{CLV}=200$, $C_{\text{offer}}=10$, and $C_{\text{contact}}=1$.\\
*CBC=CatBoost, Acc.=Accuracy, TDL=Top-Decile Lift}

\end{flushleft}
\end{table}

\subsection{Maven Dataset Analysis}

\begin{figure*}[!ht]
    \centering        
        \subfloat[Kaplan-Meier estimated retention curve \label{fig:retention2}]
        {\includegraphics[height=50mm, width=65mm]{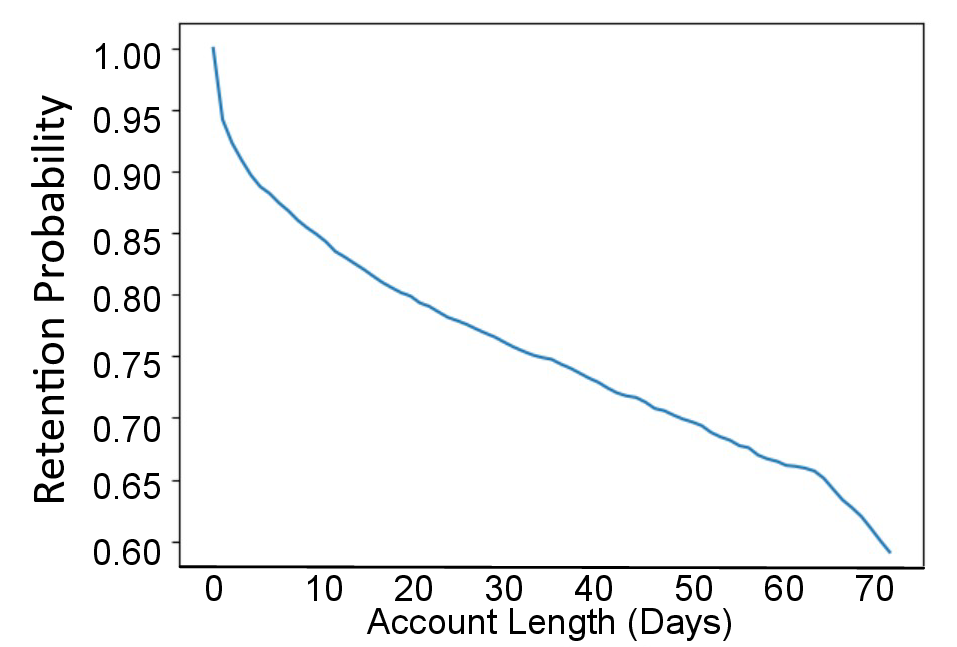}} \hfill
        \subfloat[Lift Index chart using EBM \label{fig:lift-maven}]
        {\includegraphics[height=50mm, width=65mm]{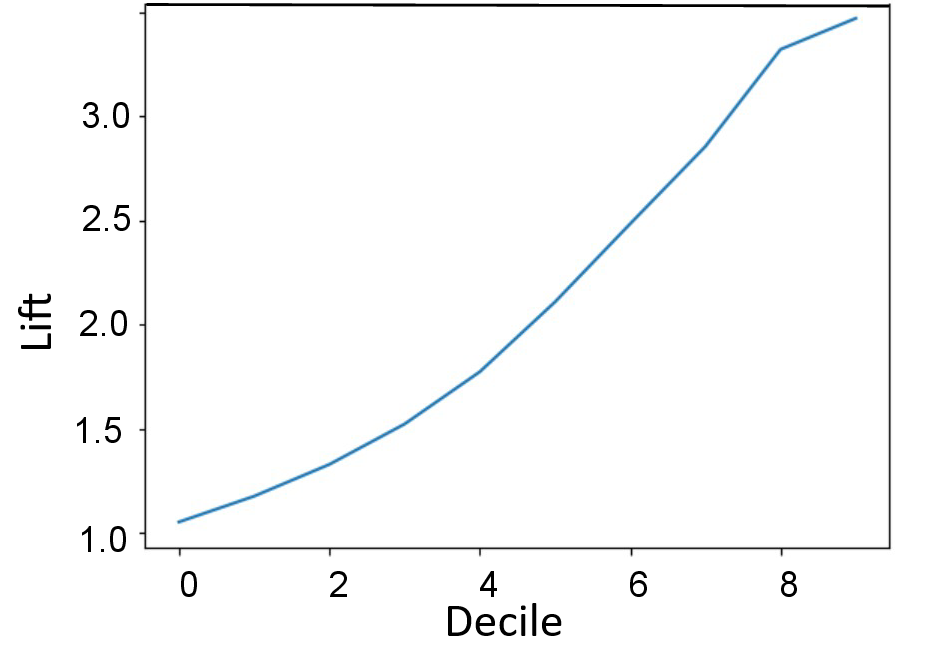}}\hfill
        \subfloat[ROC curves (tuned on \emph{e-Profits}) \label{fig:AUC-Maven}]
        {\includegraphics[height=50mm, width=75mm]{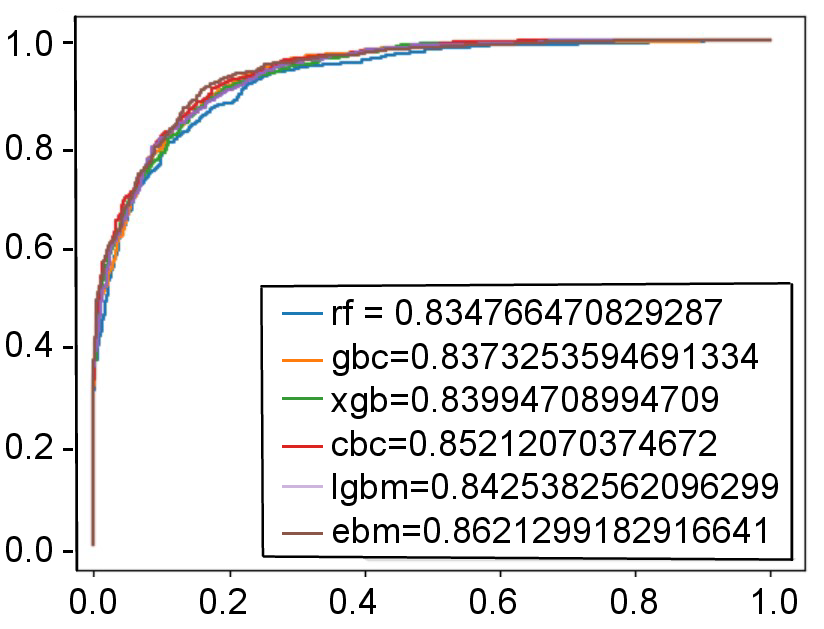}}
    \caption{Overview of Maven dataset analysis: retention estimation, churn prioritisation, and classification performance.}
    \label{fig:maven-overview}
\end{figure*}

We now turn to the Maven Telecom dataset, which features a richer set of customer attributes and a slightly different class distribution compared to IBM. Customer retention probabilities are estimated using Kaplan-Meier survival analysis. The ARR estimated from the training split of the Maven dataset is approximately 0.75. As with IBM, the Kaplan-Meier curve is fitted only on the training data and then applied unchanged during test-set evaluation. Figure~\ref{fig:maven-overview} summarises the retention estimates, lift behaviour, and classification performance. The top-decile lift reaches 3.46, indicating effective risk segmentation. Consistent with the IBM dataset, the ROC curve shows that EBM attains the highest AUC.

\subsubsection{Classifier Performance across Metrics}

Classifier performance on the Maven dataset mirrors many of the trends observed for the IBM dataset. Full results are reported in Table~\ref{table:MAVEN}, comparing five tuning strategies and evaluating both full-population and top-segment \emph{e-Profits} under ARR and TRR.

\begin{figure}
    \centering
    \includegraphics[height=75mm, width=100mm]{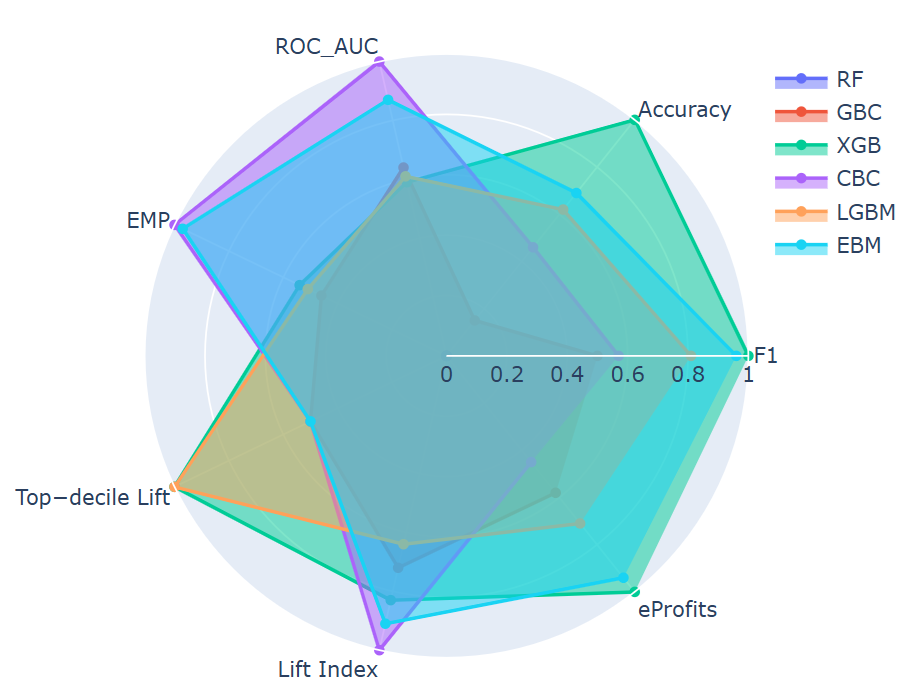}
    \caption{Radar graph for Maven dataset optimised for e-Profits on full population under TRR (axes show normalised scales for AUC, F1, Accuracy, EMP, and e-Profits).}
    \label{fig:Radar-Maven}
\end{figure}

Models optimised for AUC perform strongly on traditional predictive metrics; however, Random Forest achieves the highest \emph{e-Profits} under AUC tuning, indicating that strong ranking performance does not necessarily translate into maximum profitability. When models are tuned directly on \emph{e-Profits} (Figure~\ref{fig:Radar-Maven}), XGB delivers the highest total (full-population) profitability, while LightGBM captures the highest returns in the top 20\% customer segment.

More generally, XGB achieves the highest full-population \emph{e-Profits} when tuned on Accuracy, F1-score, and \emph{e-Profits}, and ranks second under EMP tuning, where EBM attains a slightly higher profit. In contrast, LightGBM consistently outperforms other classifiers on top-segment \emph{e-Profits} when tuned on F1-score, EMP, and \emph{e-Profits}, and ranks second when optimised for Accuracy. Consistent with the IBM dataset, directly optimising models for \emph{e-Profits} yields the highest overall profitability in the Maven dataset while preserving strong AUC and F1-score performance.

\begin{figure}
    \centering
    \includegraphics[height=75mm, width=100mm]{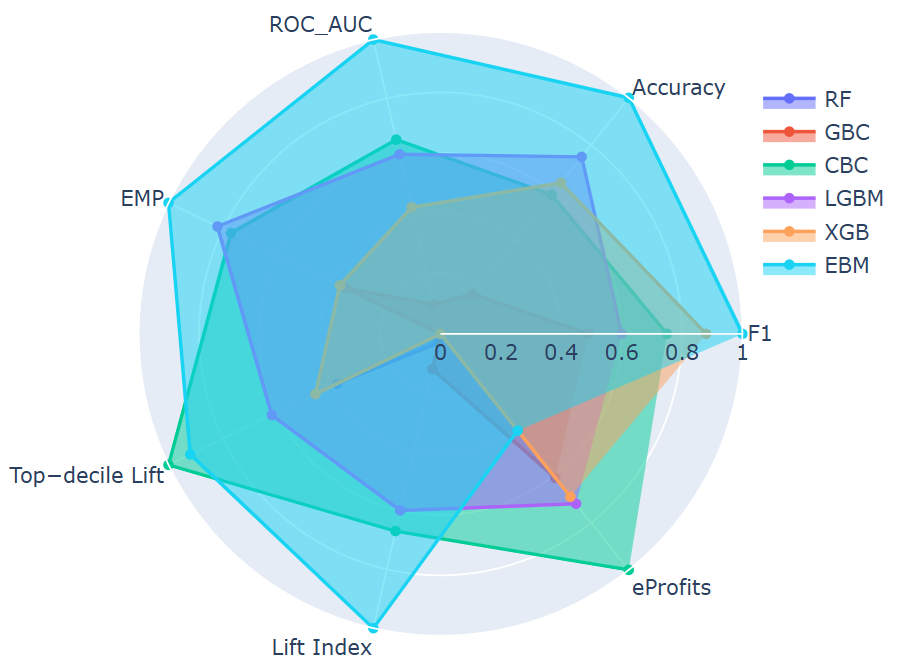}
    \caption{Radar graph for the IBM dataset optimised for e-Profits on full population under TRR (axes show normalised scales for AUC, F1, Accuracy, EMP, and e-Profits)}

    \label{fig:Radar-IBM}
\end{figure}

The radar plots shown by Figures \ref{fig:Radar-IBM} and \ref{fig:Radar-Maven} ((Higher values indicate better performance) further illustrate that no single classifier dominates across both predictive and profit-based metrics. Together, these results reinforce \emph{e-Profits} as a practical and fine-grained evaluation criterion for selecting models that maximise business value under different targeting objectives.

\begin{table}[!htbp]
\centering
\caption{\review{Classifiers' performance on the Maven dataset under AUC, Accuracy, F1-score, EMP \emph{e-Profits}, respectively.}}
\label{table:MAVEN}
\setlength{\tabcolsep}{3.2pt}
\fontsize{8pt}{12pt}\selectfont
\renewcommand{\arraystretch}{0.95}

\begin{tabular}{p{0.8cm} p{0.7cm} p{0.7cm} p{0.7cm} p{0.8cm} p{0.85cm}p{1.1cm} p{1.1cm} p{1.1cm} p{1.1cm} p{1.1cm} p{1.1cm}} 
\hline
\textbf{Name} & \textbf{F1 (\%)} & \textbf{Acc. (\%)} & \textbf{AUC (\%)} & \textbf{TDL} & \textbf{Lift Index} &
\textbf{e-Prof (0.2, 0.3, ARR)} & \textbf{e-Prof (0.2, 0.3, TRR)} &
\textbf{e-Prof (1.0, 0.3, ARR)} & \textbf{e-Prof (1.0, 0.3, TRR)} &
\textbf{EMP (0.2, 200, 10, 1)} & \textbf{EMP (1.0, 200, 10, 1)} \\
\hline

\multicolumn{12}{c}{\textbf{Tuned on AUC}} \\
\hline
RF   & 73.41 & 86.44 & 92.66 & 3.4156 & 0.8518 & \textbf{271168} & \textbf{286715} & \textbf{969408} & \textbf{996582} & 47.18 & 12.91 \\
GBC  & 68.48 & 82.90 & 91.87 & 3.4156 & 0.8460 & 67189  & 80966  & 530295 & 381167 & 46.16 & 12.77 \\
XGB  & 73.94 & 80.22 & 92.07 & 3.4508 & 0.8470 & 0      & 0      & 320168 & 140032 & 45.14 & 12.82 \\
CBC  & 72.59 & 84.37 & 92.21 & 3.3980 & 0.8475 & 111043 & 137715 & 742668 & 596238 & 45.48 & 12.76 \\
LGBM & 70.20 & 85.79 & 93.56 & \textbf{3.4684} & 0.8582 & 171384 & 213167 & 835585 & 726076 & 47.52 & 13.31 \\
EBM  & \textbf{76.11} & \textbf{87.20} & \textbf{94.16} & \textbf{3.4684} & \textbf{0.8607} & 223571 & 239644 & 969120 & 880083 & \textbf{47.72} & \textbf{13.42} \\
\hline

\multicolumn{12}{c}{\textbf{Tuned on Accuracy}} \\
\hline
RF   & 73.26 & 86.60 & 92.99 & 3.4156 & 0.8532 & 225359 & 289009 & 952522 & 868474 & 47.35 & 13.14 \\
GBC  & 75.00 & 86.75 & 93.78 & 3.4508 & 0.8589 & 259017 & 321128 & 1036960 & 966037 & 48.08 & 13.25 \\
XGB  & 75.41 & 87.00 & 94.12 & \textbf{3.4684} & 0.8612 & \textbf{268458} & \textbf{337911} & 1034441 & 958907 & \textbf{48.37} & 13.33 \\
CBC  & 75.19 & 86.95 & \textbf{94.31} & 3.4508 & 0.8616 & 244202 & 309784 & 1023033 & 948466 & 48.03 & 13.44 \\
LGBM & 73.74 & 86.60 & 94.07 & 3.4684 & 0.8611 & 217283 & 276479 & 967808 & 883372 & 48.03 & 13.44 \\
EBM  & \textbf{76.33} & \textbf{87.35} & 94.30 & 3.4683 & \textbf{0.8626} & 250647 & 316490 & \textbf{1072989} & \textbf{1003393} & 47.69 & \textbf{13.50} \\
\hline

\multicolumn{12}{c}{\textbf{Tuned on F1-score}} \\
\hline
RF   & 73.71 & 86.65 & 92.84 & 3.4332 & 0.8521 & 219722 & 279940 & 974823 & 887795 & 47.18 & 13.09 \\
GBC  & 75.00 & 86.75 & 93.78 & 3.4508 & 0.8589 & 259017 & 321128 & 1036960 & 966037 & 48.08 & 13.25 \\
XGB  & 75.41 & \textbf{87.10} & 94.12 & \textbf{3.4684} & \textbf{0.8612} & 250647 & 316490 & 1034441 & 958907 & 48.37 & 13.33 \\
CBC  & \textbf{76.19} & 86.95 & \textbf{94.31} & 3.4508 & 0.8616 & 244202 & 309784 & 1023033 & 948466 & 48.03 & \textbf{13.44} \\
LGBM & 75.88 & \textbf{87.10} & 93.84 & 3.4508 & 0.8598 & \textbf{268075} & \textbf{342257} & \textbf{1094763} & \textbf{996903} & 47.18 & 13.37\\
EBM  & \textbf{76.19} & \textbf{87.10} & 94.12 & 3.4508 & 0.8607 & 265768 & 333044 & 1075533 & 944924 & \textbf{48.25} & 13.42 \\
\hline

\multicolumn{12}{c}{\textbf{Tuned on EMP}} \\
\hline
RF   & 73.51 & 86.55 & 92.87 & 3.4332 & 0.8526 & 217001 & 278956 & 955944 & 867898 & 47.26 & 13.12 \\
GBC  & 74.83 & 86.60 & 93.67 & 3.4508 & 0.8596 & 232111 & 296995 & 1026399 & 956475 & 47.35 & 13.29 \\
XGB  & 75.74 & 87.20 & \textbf{94.21} & \textbf{3.4684} & \textbf{0.8623} & 263369 & 335457 & 1045890 & 980577 & 47.86 & 13.39 \\
CBC  & \textbf{76.22} & \textbf{87.41} & 94.15 & 3.4508 & 0.8611 & 249991 & 315433 & 1056558 & 982833 & \textbf{48.54} & 13.39 \\
LGBM & 75.88 & 87.10 & 93.84 & 3.4508 & 0.8598 & \textbf{268075} & \textbf{342257} & \textbf{1054763} & \textbf{996903} & 47.18 & 13.37 \\
EBM  & 75.50 & 86.90 & 94.08 & 3.4684 & 0.8612 & 258579 & 319910 & 1052440 & 971231 & 47.52 & \textbf{13.43} \\
\hline

\multicolumn{12}{c}{\textbf{Tuned on \emph{e-Profits} (under TRR on full population)}} \\
\hline
RF   & 73.71 & 86.65 & 92.84 & 3.4332 & 0.8521 & 219722 & 279940 & 994823 & 987795 & 47.18 & 13.09 \\
GBC  & 75.00 & 86.75 & 93.78 & 3.4508 & 0.8589 & 259017 & 321128 & 1066960 & 1046037 & 48.08 & 13.25 \\
XGB  & \textbf{76.30} & \textbf{87.30} & 93.70 & \textbf{3.4684} & 0.8600 & 231403 & 294114 & \textbf{1092216} & \textbf{1063393} & \textbf{48.54} & 13.28 \\
CBC  & 75.19 & 86.95 & \textbf{94.31} & 3.4508 & \textbf{0.8616} & 244202 & 309784 & 1053033 & 1048466 & 48.03 & \textbf{13.44} \\
LGBM & 75.80 & 87.05 & 93.73 & 3.4684 & 0.8582 & \textbf{267352} & \textbf{340635} & 1080897 & 1025717 & 48.03 & 13.27 \\
EBM  & 76.19 & 87.10 & 94.12 & 3.4508 & 0.8607 & 265768 & 333044 & 1075533 & 1004924 & 48.25 & 13.43 \\

\hline
\end{tabular}
\begin{flushleft}
\footnotesize{\emph{e-Profits} are reported as $e\text{-}Prof(\eta,M,\mathrm{RR})$, where $\eta\in\{0.2,1.0\}$ denotes the evaluated customer fraction (top 20\% or full population), $M=0.3$ is the profit margin used in Eq.~\ref{eq:clv}, and $\mathrm{RR}\in\{\mathrm{ARR},\mathrm{TRR}\}$ specifies whether average or tenure-based retention probabilities are used. EMP values are computed as $\mathrm{EMP}(\eta,\mathrm{CLV},C_{\text{offer}},C_{\text{contact}})$ with $\mathrm{CLV}=200$, $C_{\text{offer}}=10$, and $C_{\text{contact}}=1$.\\
*CBC=CatBoost, Acc.=Accuracy, TDL=Top-Decile Lift}
\end{flushleft}
\end{table}

\review{

\subsection{Statistical Significance, Uncertainty, and Model Re-Ranking}
\label{sec:statistical_significance}

To quantify uncertainty and assess whether observed differences are statistically meaningful, we report bootstrap confidence intervals and conduct paired non-parametric statistical tests. Our analysis focuses on \emph{e-Profits}, and all models are evaluated on held-out test sets for both the Maven and IBM datasets.

Table~\ref{tab:point_results} reports test-set performance for the best-performing models selected under different tuning objectives. Predictive metrics such as AUC, F1-score, and Accuracy remain broadly comparable across tuning policies, with differences typically within one to two percentage points. In contrast, \emph{e-Profits} varies substantially depending on the tuning objective. Importantly, selecting models using \emph{e-Profits} does not degrade predictive performance on either dataset, while consistently yielding the highest \emph{e-Profits}. This demonstrates that conventional classification metrics are insufficient proxies for downstream business value and that profit-aware evaluation provides complementary and decision-relevant information.

To assess estimation variability, we perform bootstrap resampling of the test sets (500 iterations) and report bootstrap means and 95\% confidence intervals for expected profit in Table~\ref{tab:bootstrap_ci}. Predictive metrics exhibit relatively narrow confidence intervals, indicating stable estimation, whereas profit-based metrics display wider intervals, reflecting the inherently higher variability of economic outcomes. Despite this variability, models selected under \emph{e-Profits} consistently achieve the highest mean profit with confidence intervals that dominate those of models selected using traditional objectives, indicating that the observed profit gains are not attributable to sampling noise.

\subsubsection{Statistical Significance of Profit Differences}

To formally test whether differences in expected profit are statistically significant, we conduct paired Wilcoxon signed-rank tests across identical model families evaluated on the same test sets. These tests operate on paired profit differences and assess whether profit gains under \emph{e-Profits} are systematic rather than random. Results are summarized in Table~\ref{tab:wilcoxon}. Across both the Maven and IBM datasets, models selected using \emph{e-Profits} significantly outperform those tuned for AUC and Accuracy in terms of full-population \emph{e-Profits} (\(p \ll 0.001\)). The magnitude of these gains is substantial on both datasets, confirming that profit improvements under \emph{e-Profits} are statistically significant and consistent across model families.

\subsubsection{Model Re-Ranking Under Profit-Based Evaluation}

While predictive metrics often rank models similarly, evaluating models using expected profit induces systematic reordering. To quantify this effect, we compute Spearman’s rank correlation between AUC-based rankings and profit-based rankings. Statistical significance is assessed via paired Wilcoxon signed-rank tests on rank differences.

Results are reported in Table~\ref{tab:rerank} showing agreement between AUC-based and profit-based rankings is moderate rather than perfect on both the datasets. On the Maven dataset, Spearman’s rank correlation is approximately $\rho=0.49$ for full-population e-Profits and $\rho=0.45$ for top-20\% targeting. On IBM, rank agreement is slightly lower but comparable, with $\rho\approx0.43$ for full-population profit and $\rho\approx0.39$ for top-20\% profit. In all cases, paired Wilcoxon signed-rank tests on rank differences strongly reject perfect concordance ($p<10^{-23}$ for Maven and $p<10^{-18}$ for IBM), confirming that profit-based evaluation induces systematic and statistically significant re-ranking rather than random variation. These results indicate that predictive performance metrics capture only part of the information relevant for economic decision-making, while profit-aware evaluation provides complementary guidance, particularly for targeted retention strategies.

Taken together, these results demonstrate that business-aware evaluation using \emph{e-Profits} leads to statistically justified model selection decisions and materially different rankings than traditional predictive metrics. These effects are consistent across datasets with differing characteristics, reinforcing the importance of profit-aware evaluation in applied decision-making contexts.

\begin{table}[!htbp]
\centering
\caption{Test-set performance of best-performing models selected under different tuning objectives for the Maven and IBM datasets.}
\label{tab:point_results}

\begin{tabular}{p{1.1cm} p{1.2cm} p{1.2cm} p{1.4cm} p{1.2cm} p{1.3cm} p{3.2cm}}
\hline
\textbf{Dataset} & \textbf{Policy} & \textbf{Model} & \textbf{AUC (\%)} & \textbf{F1 (\%)} & \textbf{Acc. (\%)} & \textbf{e-Profits (Full TRR)} \\
\hline
\multirow{5}{*}{\textbf{Maven}} 
&AUC              & RF   & 92.63 & 73.41 & 86.44 & 996582 \\
&Accuracy         & EBM  & 94.30 & 76.33 & 87.35 & 1003393 \\
&F1               & LGBM & 93.84 & 75.88 & 87.10 & 996903 \\
&EMP              & LGBM & 93.84 & 75.88 & 87.10 & 996903 \\
&\emph{e-Profits} & XGB  & 93.70 & 76.30 & 87.30 & \textbf{1063393} \\
\hline
\multirow{5}{*}{\textbf{IBM}} 
&AUC              & RF   & 83.77 & 57.20 & 79.18 & 9104621 \\
&Accuracy         & XGB  & 85.48 & 59.32 & 80.27 & 9953379 \\
&F1               & XGB  & 85.58 & 59.51 & 80.36 & 9953379 \\
&EMP              & LGBM & 84.68 & 56.87 & 79.60 & 9829970 \\
&\emph{e-Profits} & XGB  & 83.82 & 58.47 & 79.22 & \textbf{10991460} \\
\hline
\end{tabular}
\begin{flushleft}
*For each tuning policy, the \emph{model} column reports the classifier that achieves the highest \emph{e-Profits} on the held-out test set among all models trained under that policy. The reported AUC, F1-score, and Accuracy values correspond to this profit-maximising model and are not necessarily the maximum values achievable under the tuning objective.

\end{flushleft}
\end{table}

\begin{table}[!htbp]
\centering
\caption{Bootstrap mean and 95\% confidence intervals (500 resamples) for e-Profits (Full TRR) on Maven and IBM dataset.}
\label{tab:bootstrap_ci}
\begin{tabular}{p{2.0cm} p{2.0cm} p{2.0cm} p{2.0cm} p{2.0cm}}
\hline
\textbf{Dataset} & \textbf{Policy} & \textbf{Mean} & \textbf{CI Low} & \textbf{CI High} \\
\hline
\multirow{3}{*}{\textbf{Maven}} 
& AUC       & 996925   & 862389   & 1127743 \\
& Accuracy  & 1003816 & 871303   & 1137394 \\
& e-Profits & \textbf{1061128} & 875735 & 1129249 \\
\hline
\multirow{3}{*}{\textbf{IBM}} 
& AUC       & 9103018 & 8109360 & 10180288 \\
& Accuracy  & 9952973 & 8923399 & 11018421 \\
& e-Profits & \textbf{10913261} & 8945986 & 11217734 \\
\hline
\end{tabular}
\end{table}

\begin{table}[!htbp]
\centering
\caption{Paired Wilcoxon signed-rank tests comparing bootstrap mean differences in full-population TRR-based \emph{e-Profits} across identical model families under different tuning objectives.}
\label{tab:wilcoxon}
\begin{tabular}{p{1.2cm} p{1.2cm} p{1.4cm} p{2.8cm} p{1.4cm} p{1.4cm}}
\hline
\textbf{Dataset} & \textbf{Policy A} & \textbf{Policy B} & \textbf{Metric} & \textbf{Mean Diff} & \textbf{$p$-value} \\
\hline
\multirow{2}{*}{Maven}
&e-Profits & AUC      & e-Profits (Full TRR) & +64203 & $\ll 0.001$ \\
&e-Profits & Accuracy & e-Profits (Full TRR) & +57312  & $\ll 0.001$ \\
\hline
\multirow{2}{*}{IBM}
&e-Profits & AUC      & e-Profits (Full TRR) & +1810243 & $\ll 0.001$ \\
&e-Profits & Accuracy & e-Profits (Full TRR) & +960288 & $\ll 0.001$ \\
\hline
\end{tabular}
\end{table}

\begin{table}[!htbp] 
\centering 
\caption{Agreement between AUC-based ranking and profit-based ranking, measured by Spearman’s $\rho$.}
\label{tab:rerank} 
 \begin{tabular}{p{1.2cm} p{3.0cm} p{3.4cm} p{2.0cm}} 
 \hline 
 \textbf{Dataset} & \textbf{Profit metric} & \textbf{Spearman $\rho$ (95\% CI)} & \textbf{$p$-value} \\ 
 \hline 
 \multirow{2}{*}{Maven}
    & eProfits (Full TRR) & $\approx 0.49$ & $<10^{-24}$ \\ 
    & eProfits (Top20 TRR)& $\approx 0.45$ & $<10^{-23}$ \\ 
    \hline 
    \multirow{2}{*}{IBM}
    & eProfits (Full TRR) & $\approx 0.43$ & $1.50\times 10^{-18}$ \\ 
    & eProfits (Top20 TRR)& $\approx 0.39$ & $7.07\times 10^{-19}$ \\ \hline 
 \end{tabular} 
\end{table}

}

\section{Discussion and Implications}
\label{sec:discussion}

Our findings across the IBM and Maven datasets reveal consistent patterns that underscore the practical value of using \emph{e-Profits} as an evaluation metric for churn prediction models. Traditional performance metrics such as AUC and F1-score do not consistently identify the most financially effective models. For instance, classifiers such as LightGBM and CatBoost, which do not always top the rankings under Accuracy or AUC, emerge as highly profitable when evaluated using \emph{e-Profits}. This reordering highlights a critical discrepancy between predictive correctness and financial value, underscoring the limitations of relying solely on statistical metrics for real-world business decisions.

Moreover, \emph{e-Profits} proves especially effective in identifying models that excel within high-value customer segments. Across both datasets, models such as LightGBM and XGBoost frequently deliver the highest profitability when evaluation is restricted to the top 20\% of customers ranked by expected profit. These models are not necessarily preferred under traditional metrics, reinforcing the importance of evaluation criteria that prioritise return on investment (ROI) over aggregate predictive accuracy. The ability to tailor model evaluation toward strategic, high-impact interventions makes \emph{e-Profits} particularly suitable for resource-constrained retention campaigns.

These results confirm that explicitly coupling churn risk with heterogeneous customer value and intervention cost leads to systematically different model rankings than accuracy-based evaluation. Models with similar AUC or F1-score can differ substantially in how they prioritise high-CLV customers or avoid costly false positives. By translating predictions into per-customer economic outcomes, \emph{e-Profits} amplifies these differences and reveals profit-relevant distinctions that remain invisible to purely statistical metrics.

A further insight from our analysis concerns the limited utility of commonly used lift-based measures in reflecting actual business outcomes. While lift remains a popular heuristic in industry, it does not account for heterogeneity in customer value or retention cost. For example, Table~\ref{table:IBM} illustrates scenarios in which models with strong lift scores nevertheless deliver zero \emph{e-Profits}, indicating that churn likelihood alone is an insufficient criterion for profitable targeting. By jointly considering churn risk, revenue potential, and intervention cost, \emph{e-Profits} provides a more holistic and economically meaningful basis for model selection.

From an implementation perspective, \emph{e-Profits} requires reliable estimation of customer lifetime value and retention probabilities, which in turn depends on the availability of revenue, tenure, and churn-timing data. In addition, profitability outcomes are sensitive to cost parameters such as contact and offer costs, which must be calibrated to organisational policies and campaign constraints. While our experiments adopt domain-informed heuristics, these parameters can be readily adapted to reflect business-specific assumptions. Our current formulation assumes a binary intervention decision; extending the framework to support multi-tier or continuous intervention strategies represents a natural extension.

Our results do not imply that traditional metrics such as AUC or F1-score are obsolete. These metrics remain appropriate when the primary objective is ranking quality or classification accuracy under uniform cost assumptions. However, when customer value is heterogeneous, intervention costs scale with CLV, or budgets are constrained, profit-aware evaluation becomes essential. In such settings, \emph{e-Profits} offers a more decision-relevant criterion for model selection.

By explicitly aligning model evaluation with customer value, retention probability, and intervention cost, \emph{e-Profits} provides a decision-oriented criterion that better reflects operational objectives. While profit-based alternatives such as EMP move in this direction, their reliance on fixed population-level assumptions limits their effectiveness in heterogeneous customer bases.

\review{
\subsection{Managerial Implications, Applicability}

From a managerial perspective, our findings indicate that selecting churn models solely based on predictive metrics such as AUC or F1-score can lead to systematically suboptimal retention strategies, particularly under budget constraints. In practice, organisations rarely intervene on all predicted churners; instead, retention campaigns must allocate limited resources across customers with heterogeneous value. As a result, model evaluation criteria that ignore customer value and intervention costs can misguide operational decision-making.

Deploying models evaluated with \emph{e-Profits} enables firms to (i) select different models for mass targeting versus high-value, budget-limited campaigns, (ii) prioritise models that reduce costly false positives among low-value customers, and (iii) explicitly align model evaluation with campaign economics rather than abstract predictive performance. This allows data science teams to justify model deployment decisions in financial terms that are directly interpretable and actionable for marketing, CRM, and customer success stakeholders. Although absolute profit values naturally scale with the choice of cost parameters, we empirically observed that relative model rankings remained stable across reasonable parameter ranges (e.g., $\mathrm{CPO} \in [0.05, 0.2]$). This observation motivated our decision to focus on ranking robustness rather than exhaustive sensitivity grids in the present study.

Although our empirical evaluation uses telecom datasets, the \emph{e-Profits} framework is not tied to telecom-specific features. In practice, customer lifetime value and retention can be derived from basic billing and tenure information (e.g., monthly revenue and time-to-churn or subscription length), while intervention costs are specified through business assumptions rather than inferred directly from the data. This makes \emph{e-Profits} applicable to other subscription-based domains such as banking or SaaS, provided that comparable revenue and tenure signals are available and an operational definition of churn exists.

The main challenges are therefore practical rather than methodological. In many non-telecom settings, suitable customer-level data are not publicly available for experimentation, and churn dynamics (e.g., soft churn or account dormancy) may require domain-specific calibration of value and cost components. Nevertheless, these challenges concern data availability and business calibration rather than limitations of the evaluation framework itself.

While our experiments use fixed cost and margin values, \emph{e-Profits} is inherently parametric. Varying these parameters is expected to rescale profits and shift targeting thresholds without altering the underlying decision logic, allowing organisations to calibrate the metric to their own cost structures. 

\subsection{Limitations and Future Directions}

This study positions \emph{e-Profits} as a decision-oriented evaluation and model-selection metric rather than a causal treatment-effect estimator. We assume that intervening on a true churner successfully prevents churn and do not explicitly model heterogeneous responsiveness to retention actions. As a result, \emph{e-Profits} evaluates the profitability of a targeting policy under explicit cost and retention assumptions rather than estimating incremental profit relative to a no-intervention counterfactual. Integrating causal uplift or treatment-effect modelling to estimate customer-specific intervention effectiveness therefore represents an important direction for future work.

A further limitation concerns the estimation of retention probabilities using Kaplan-Meier survival analysis. While Kaplan-Meier is non-parametric, transparent, and computationally lightweight, it assumes non-informative censoring and captures retention heterogeneity only through customer tenure using a single population-level survival curve. In settings where churn dynamics depend strongly on customer covariates or evolve over time, this assumption may be restrictive. Future work may therefore explore more expressive survival models to capture heterogeneity beyond tenure, including parametric approaches (e.g., Weibull) and covariate-aware models such as Cox proportional hazards with time-varying covariates. Marketing-oriented parametric churn models such as Beta-Geometric / BG-NBD formulations also represent a promising alternative, offering smoother and more stable retention estimates in sparse or long-tenure regions.

Finally, while our experiments focus on telecom datasets, the \emph{e-Profits} framework is not domain-specific. Applying the metric in other subscription-based contexts (e.g., banking or SaaS) would require domain-calibrated definitions of churn, revenue, and intervention costs. Investigating domain transfer, parameter calibration, and extensions to multi-tier or continuous intervention strategies represents an important avenue for future research.

}

\section{Conclusion}
\label{sec:conclusion}

Traditional churn prediction models are commonly evaluated using statistical performance metrics such as Accuracy, AUC, and F1-score. While these metrics are useful for assessing classification performance, they often fail to reflect the financial consequences of misclassification, particularly in profit-sensitive applications such as customer retention.

In this paper, we introduced \emph{e-Profits}, a business-aligned evaluation metric that integrates customer-specific lifetime value, intervention costs, and retention probabilities derived from Kaplan-Meier survival analysis. Unlike traditional metrics and profit-based alternatives such as EMP, \emph{e-Profits} evaluates models based on individualised cost–revenue dynamics, enabling closer alignment between predictive analytics and financial outcomes. Empirical results on two benchmark datasets show that \emph{e-Profits} induces profit-relevant model re-ranking: models that are suboptimal under Accuracy or AUC often emerge as financially superior, particularly when evaluation focuses on high-value customer segments. Overall, the consistent divergence between full-population and top-segment profitability has direct implications for retention campaign design.

Beyond empirical performance, \emph{e-Profits} is designed for practical deployment. \emph{e-Profits} provides a more decision-relevant basis for model selection. Its plug-in formulation allows it to be applied to any churn classifier using predicted probabilities, without modifying the learning algorithm, and it integrates naturally into standard evaluation and model-selection workflows such as cross-validated tuning. The transparent, per-customer profit decomposition further facilitates communication between technical teams and business stakeholders. Overall, these results demonstrate that traditional predictive metrics are insufficient proxies for business value in churn prediction, and that profit-aware evaluation using \emph{e-Profits} provides a more decision-relevant basis for model selection.

\bmhead{Acknowledgments}
This publication has emanated from research conducted with the support of Research Ireland under award no. GOIPG/2021/1354, and with the financial support of Research Ireland under grant no. 13/RC/2106\_P2 at the ADAPT Research Centre at Technological University Dublin.

\section*{Declaration} 
The authors declare that they have no known competing financial interests or personal relationships that could have appeared to influence the work reported in this paper. 

\section*{Funding}
This publication has emanated from research conducted with the support of Research Ireland under award no. GOIPG/2021/1354, and with the financial support of Research Ireland under grant no. 13/RC/2106\_P2 at the ADAPT Research Centre at Technological University Dublin.

\section*{Ethics approval and consent to participate}
Not applicable

\section*{Data availability} 
\textbf{IBM Telco Dataset} is available at \url{https://www.kaggle.com/datasets/blastchar/telco-customer-churn/data} and \textbf{Maven Telecom Dataset} is available at \url{https://www.kaggle.com/datasets/johnp47/maven-churn-dataset}. 

\section*{Materials availability}
Not applicable

\section*{Code availability} 
All source code is available at: \url{https://github.com/Awaismanzoor/eprofits}

\section*{Author contribution}
\textbf{Awais Manzoor:} Conceptualization, Data curation, Methodology and Experimental setup, Software, Formal analysis, Investigation, Validation, Writing – original draft, Writing – review \& editing \textbf{M. Atif Qureshi:} Conceptualization, Data curation, Methodology and Experimental setup, Software, Formal analysis, Investigation, Validation, Writing – review \& editing, Supervision \textbf{Etain Kidney:} Supervision, Writing – review \& editing \textbf{Luca Longo:} Validation

\begin{appendices}
\review{
\section{Toy Example: \emph{e-Profits} Calculation}
\label{sec:toy_example}

\subsection{\emph{e-Profits} Calculation Using TRR (conditional one-period retention)}
To illustrate how \emph{e-Profits} evaluates profitability at the customer level, we present a toy example of five customers with known churn labels $y_i$, predictions $\hat{y}_i$, and financial attributes. Following the targeting policy in Eq.~\ref{eq:eprofits}, only customers predicted as churners ($\hat{y}_i = 1$) are considered for intervention.

In this example, the tenure-based retention probability (TRR) is interpreted as a conditional one-period retention probability. Let $S(\cdot)$ denote the Kaplan-Meier survival function estimated from churn durations. Given a customer with current tenure $T_i$, the conditional retention over a horizon $\Delta$ is
\[
r_i^{\text{TRR}}(\Delta) = \Pr(T > T_i+\Delta \mid T>T_i)=\frac{S(T_i+\Delta)}{S(T_i)}.
\]
For simplicity, we use $\Delta=1$ (one period, e.g., one month). The retention values shown in Table~\ref{tab:toydata} should be understood as $r_i^{\text{TRR}}(1)$ (i.e., one-period conditional retention), obtained by evaluating the fitted Kaplan-Meier curve.

\begin{table}[ht]
\centering
\caption{Example customer data for \emph{e-Profits} illustration (TRR values are conditional one-period retention probabilities)}
\label{tab:toydata}
\begin{tabular}{p{3.0pc}p{3.7pc}p{3.7pc}p{4.9pc}p{6.6pc}p{6.4pc}}
\hline
\textbf{Customer} & \textbf{$y_i$ (True)} & \textbf{$\hat{y}_i$ (Pred)} & \textbf{$R_i$ (Revenue)} &
\textbf{$r_i^{\text{TRR}}(1)$ (Cond.\ retention)} & \textbf{Account Length $T_i$} \\
\hline
A & 1 & 1 & 100 & 0.80 & 120 \\
B & 0 & 1 & 80  & 0.90 & 50 \\
C & 1 & 0 & 120 & 0.75 & 180 \\
D & 0 & 0 & 90  & 0.85 & 90 \\
E & 1 & 1 & 150 & 0.70 & 200 \\
\hline
\end{tabular}
\end{table}

Let us assume the profit margin $M = 0.3$, the cost-per-offer ratio $\text{CPO} = 0.1$, and contact cost parameters $c_0 = 5$, $c_1 = 0.3$\footnote{The contact cost is determined as follows: $c_0$ represents the minimum cost incurred regardless of CLV (e.g., email costs, phone contact). This minimum cost is compared to a CLV-based proportionate contact cost ($c_1$ times the offer cost), and the greater value is selected.}. Under TRR, CLV is computed as:
\[
\text{CLV}_i^{\text{TRR}} = \frac{R_i \cdot M}{1 - \min\!\left(r_i^{\text{TRR}}(1),0.995\right)}.
\]
The intervention costs for each predicted churner are:
\[
C_{\text{offer}, i} = \text{CPO} \cdot \text{CLV}_i, \quad
C_{\text{contact}, i} = \max(c_0,\, c_1 \cdot C_{\text{offer}, i}).
\]

\textbf{Customer A (TP):}
\[
\text{CLV}_A = \frac{100 \cdot 0.3}{1 - 0.80} = 150,\quad
C_{\text{offer}} = 15,\quad
C_{\text{contact}} = \max(5, 4.5)=5
\]
\[
\Pi_A = 150 - 15 - 5 = \mathbf{130}
\]

\textbf{Customer B (FP):}
\[
\text{CLV}_B = \frac{80 \cdot 0.3}{1 - 0.90} = 240,\quad
C_{\text{offer}} = 24,\quad
C_{\text{contact}} = \max(5, 7.2)=7.2
\]
\[
\Pi_B = -24 - 7.2 = \mathbf{-31.2}
\]

\textbf{Customer C and D:}
\[
\hat{y}_C = \hat{y}_D = 0 \Rightarrow \Pi_C = \Pi_D = 0
\]

\textbf{Customer E (TP):}
\[
\text{CLV}_E = \frac{150 \cdot 0.3}{1 - 0.70} = 150,\quad
C_{\text{offer}} = 15,\quad
C_{\text{contact}} = \max(5, 4.5)=5
\]
\[
\Pi_E = 150 - 15 - 5 = \mathbf{130}
\]

\textbf{Total \emph{e-Profits} (TRR-based):}
\[
\sum \Pi_i^{\text{TRR}} = 130 + (-31.2) + 0 + 0 + 130 = \mathbf{228.8}.
\]

\vspace{1mm}
\subsection{\emph{e-Profits} Calculation Using ARR (global one-period retention)}
To contrast tenure-based retention with a population-level assumption, we repeat the toy example using an Average Retention Probability (ARR). To remain consistent with the conditional TRR definition, we define ARR as the mean of the conditional one-period retention probabilities:
\[
r^{\text{ARR}}(1) = \frac{1}{N}\sum_{i=1}^{N} r_i^{\text{TRR}}(1).
\]
Using the five customers:
\[
r^{\text{ARR}}(1)=\frac{0.80+0.90+0.75+0.85+0.70}{5}=\frac{4.00}{5}=0.80.
\]
ARR-based CLV applies this global retention uniformly:
\[
\text{CLV}_i^{\text{ARR}} =
\frac{R_i \cdot M}{1 - \min\!\left(r^{\text{ARR}}(1),0.995\right)}.
\]
With $r^{\text{ARR}}(1)=0.80$, this becomes:
\[
\text{CLV}_i^{\text{ARR}} = \frac{0.3 \cdot R_i}{1 - 0.80}
= \frac{0.3 \cdot R_i}{0.2}=1.5\cdot R_i.
\]
Intervention costs are computed identically:
\[
C_{\text{offer}, i} = \text{CPO} \cdot \text{CLV}_i, \quad
C_{\text{contact}, i} = \max(c_0,\, c_1 \cdot C_{\text{offer}, i}).
\]

\textbf{Customer A (TP):}
\[
\text{CLV}_A = 1.5 \times 100 = 150,\quad
C_{\text{offer}} = 15,\quad
C_{\text{contact}} = \max(5, 4.5)=5
\]
\[
\Pi_A = 150 - 15 - 5 = \mathbf{130}
\]

\textbf{Customer B (FP):}
\[
\text{CLV}_B = 1.5 \times 80 = 120,\quad
C_{\text{offer}} = 12,\quad
C_{\text{contact}} = \max(5, 3.6)=5
\]
\[
\Pi_B = -12 - 5 = \mathbf{-17}
\]

\textbf{Customer C and D:}
\[
\hat{y}_C = \hat{y}_D = 0 \Rightarrow \Pi_C = \Pi_D = 0
\]

\textbf{Customer E (TP):}
\[
\text{CLV}_E = 1.5 \times 150 = 225,\quad
C_{\text{offer}} = 22.5,\quad
C_{\text{contact}} = \max(5, 6.75)=6.75
\]
\[
\Pi_E = 225 - 22.5 - 6.75 = \mathbf{195.75}
\]

\textbf{Total \emph{e-Profits} (ARR-based):}
\[
\sum \Pi_i^{\text{ARR}} = 130 + (-17) + 0 + 0 + 195.75 = \mathbf{308.75}.
\]

This toy example shows how \emph{e-Profits} combines prediction correctness with customer-level economics. Customers A and E contribute positively because they are true positives, while Customer B produces a loss as a false positive due to unnecessary intervention costs. The difference between TRR and ARR arises because TRR uses customer-specific conditional retention $r_i^{\text{TRR}}(1)$ whereas ARR applies a single averaged retention probability $r^{\text{ARR}}(1)$ uniformly.}
\end{appendices}

\bibliography{ref}
\vspace{6mm}

\end{document}